# Visual-Inertial RGB-D SLAM with Encoders for a Differential Wheeled Robot

Zhanghao Zhu, Yutaka Kaizu, Kenichi Furuhashi, and Kenji Imou

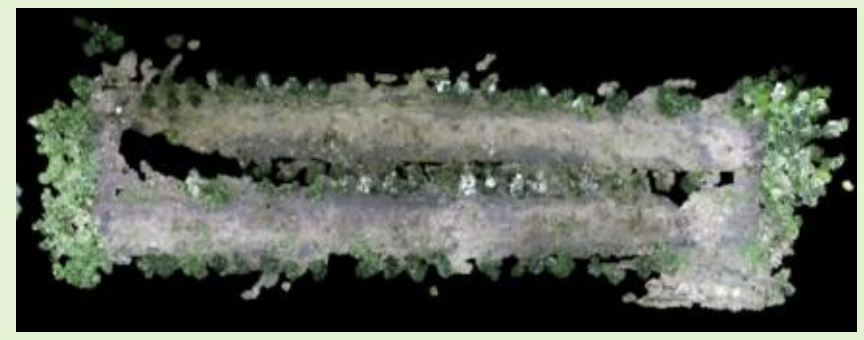

***Abstract*—Recent years have seen multiple impressive results in visual-inertial odometry (VIO) techniques, by which accurate state estimation can be achieved via the extended Kalman filter (EKF) or nonlinear optimization. However, these approaches are rarely open source, and they tend to fail in real experiments due to the temporary lack of feature points and fast motion. Therefore, in this study, we used encoders to overcome the temporary failure of purely vision-based simultaneous localization and mapping (SLAM). Here, we propose a generative measurement model for encoders and derive an expression for the maximum a posteriori (MAP) estimate and necessary Jacobians for optimization. We use our theory to present a novel tightly coupled visual-inertial encoder RGB-Depth (RGB-D) SLAM system. Tests on our system on an in-house dataset confirmed that our modeling effort led to accurate (with a root mean squared error (RMSE) of approximately 2-7 cm) and robust state estimation in real time. The source code and our dataset containing the encoder information have been published for verification.**

***Index Terms*—Encoder, Sensor fusion, Simultaneous localization and mapping, Visual-inertial odometry**

## I. Introduction

IN agriculture, some types of repetitive and hazardous work, e.g., harvesting, spraying and pruning, can be performed by mobile robots to improve the efficiency and safety [1]. However, in places where global positioning system (GPS) signals are not fully available, such as greenhouses or orchards, an extensive and costly modification of the environment, such as rail laying, is required to operate these robots. To avoid this disadvantage, robots are required to recognize the surrounding natural environment and freely run. A low-cost localization and navigation technique that uses a vision sensor with an open-loop visual odometry (VO) algorithm, GPS capabilities, and an inertial measurement unit (IMU) for vineyards was developed [2]. In greenhouses or orchards with prolific canopies, GPS is not reliable [3]; therefore, a 3D measurement of the environment is used. A map of a block in an orchard was built with inexpensive onboard sensors: lidar and wheel encoders [4].

Simultaneous localization and mapping (SLAM) technology has attracted attention in recent years. This approach can help a robot determine its location and intended destination even when a GPS signal is not available. When such a robot is placed in an unknown environment, it incrementally builds a consistent map of the environment and applies it to determine its location [5]. In the present study, SLAM is used to navigate a robot.

Monocular SLAM has an initialization problem when a differential wheeled robot lacks significant parallax [6]. Stereo SLAM must spend more time computing two images. Thus, in this study, we decided to select RGB-D SLAM for stable and real-time consideration.

A well-known RGB-D SLAM called Rgbdslamv2 was proposed [7]. The same authors also presented a thorough experimental evaluation on a publicly available benchmark dataset [8]. Based on their successful experiment, we selected Microsoft Kinect v2 as the RGB-D camera in our study.

An Open-Source SLAM System for Monocular, Stereo, and RGB-D Cameras (ORBSLAM2) has attracted the attention of researchers due to its high accuracy and real-time speed [9]. We greatly benefited from its powerful but fast loop detector [10] because we introduce pure encoder edges, which connect the states of different frames to avoid temporary camera failure; this detector helps to improve the accuracy. ORBSLAM2 runs at approximately 30 Hz on our computing facilities and obtains the most accurate result of the packages applied.

Although cameras can achieve an accuracy measured in centimeters, they tend to lose tracking in environments with few features and poor light conditions. Tracking failures arise due to fast camera motion, the unstable performance of an embedded board, or a low frame rate of capturing images. For the real implementation of SLAM in sweeping robots, for example, researchers usually choose IMU sensors to provide self-motion information, especially for recovering metric scales for monocular cameras and compensating for the effect of dynamic objects in scenes. In the present study, we used supplementary sensors such as an IMU and encoders to avoid the temporary tracking failure problems that might occur in pure vision SLAM.

Visual-inertial SLAM is currently a major research topic that

This research was supported by grants from the Project of the Bio-oriented Technology Research Advancement Institution NARO (the special scheme project on advanced research and development for next-generation technology).

Z. Zhu, Y. Kaizu, K. Furuhashi, and K. Imou are with the Graduate School of Agricultural and Life Sciences, The University of Tokyo, 1-1-1, Yayoi, Bunkyo-ku, Tokyo 113-8657, Japan (e-mail: zhuzhanghao9331@yahoo.co.jp, {kaizu, kfuruhashi, k-imou}@g.ecc.u-tokyo.ac.jp).

can be divided into two categories: loosely coupled and tightly coupled. The latter, which includes features or landmarks in the state vector, usually achieves better accuracy. The well-known tightly coupled SLAM systems include robust visual-inertial odometry, which is based on the EKF [11], Open Keyframe-based Visual-Inertial SLAM (OKVIS) [12], a Monocular Visual-Inertial System (VINS-Mono) [13] and the direct method [14]. Among these visual-inertial SLAM methods, the most accurate result for the EuRoC dataset [15] is obtained by Visual-Inertial ORBSLAM2 (VIORBSLAM) [16], which is mainly implemented on the theoretical basis of real-time IMU preintegration [17]. In the VIORBSLAM paper, a novel IMU initialization method called a divide and conquer approach was proposed to compute an initial solution of the scale, gravity direction, velocity, and biases of the gyroscope and accelerometer.

However, this IMU initialization method requires that the pure vision SLAM survive for approximately 15 s to make the bias and gravity variables observable [16]. Since the present project pursues a more robust visual SLAM method with the aim of maximizing performance in real situations, one more odometry sensor without any bias variable should be introduced. As the GPS is unreliable in some cases, assuming that the failure of the camera is temporary and supposing that the ground is piecewise flat and provides enough friction, in this project, we utilize Hall sensors of brushless motors as two wheel encoders. Especially on almost flat ground, the IMU may be omitted because it constrains the time span between consecutive keyframes to no more than 0.5 s [16], which leads to a large map size, and because the encoder has already provided enough information, the computational cost for the IMU can be eliminated.

The objective of this research is to strengthen the robustness of RGB-D ORBSLAM2 through the utilization of onboard sensors such as encoders and an IMU for a differential wheeled robot. Here, we define the robustness as follows:

- The temporary failure of pure vision SLAM is reduced due to the lack of feature points, light problems, fast motion, low frame-per-second (fps) images, or the unstable execution capability of the microcontroller unit (MCU).
- The system obtains as many trajectories and map points as possible instead of using the relocalization strategy without substantially compromising the accuracy.

The first step toward this objective is to introduce novel pure encoder edges to ORBSLAM2. We present a new flow chart in Section III. We found that although these encoder edges may resolve the temporary failure of pure vision, the accuracy will be greatly reduced due to the lack of a covariance matrix. Our second contribution is to develop a novel preintegration theory on encoders and frames for ORBSLAM2. Inspired by the IMU preintegration [17], we derive encoder preintegration and present the related bundle adjustment (BA) changes in Sections II~IV. Compared to [17], we propose a novel encoder model, which permits error between the real motion and the suggested model. We provide an exact calculation result under the piecewise flat ground assumption, while [17] derives an approximate result. Here, we refer to RGB-D ORBSLAM2 tightly coupled with both encoders and an IMU as VIEOS2, RGB-D ORBSLAM2 coupled with only encoders as VEOS2, and RGB-D ORBSLAM2 coupled with only an IMU as VIOS2. They are compatible with the stereo part, and the corresponding name will add “Stereo” as a postfix, e.g., VIOS2 Stereo. For comparison with VIORBSLAM, our implemented Monocular ORBSLAM2 coupled with an IMU is called VIOS2 Monocular. To the best of our knowledge, this is the first trial of a tightly coupled open-source visual-inertial encoder for RGB-D SLAM. We published our code at https://github.com/leavesnight/ORB_SLAM2 and the agricultural dataset with the encoder information at https://pan.baidu.com/s/1ZUZ7ZsemFtZjm5r3ve62TA(Password:237g). We share a video of our VEOS2 in comparison with ORBSLAM2 on our dataset at https://www.facebook.com/100010550718290/videos/601063854214041/.

## II. VIEOS2 PRELIMINARIES

This project derives the error expression of the encoder term and the iterative form of the covariance transfer and their Jacobians in a similar manner to [17] but uses some formulas introduced in [18]. Therefore, only some new points relevant to the introduction of the encoders are reviewed in this section.

The notions of Riemannian geometry, uncertainty description in the special orthogonal group (SO(3)), and the Gauss-Newton (GN) method on manifolds were well introduced in [17] and [18]. We also use the trust region improved GN method named the Levenberg-Marquardt (LM) method, which was implemented in g2o [19].

$$
\begin{aligned}
&(\mathbf{H}+\lambda\mathbf{I})\Delta\mathbf{x}^{*}=-\mathbf{b}\\
&\mathbf{H}_{ij}=\sum_{\mathbf{r}_{(\cdot)}}\dot{\rho}\left(\mathbf{r}_{(\cdot)}^{\mathrm{T}}\boldsymbol{\Sigma}_{(\cdot)}^{-1}\mathbf{r}_{(\cdot)}\right)\mathbf{J}_{\mathbf{r}_{(\cdot)}}^{\mathbf{x}_i\mathrm{T}}\boldsymbol{\Sigma}_{(\cdot)}^{-1}\mathbf{J}_{\mathbf{r}_{(\cdot)}}^{\mathbf{x}_j} \qquad (1)\\
&\mathbf{b}_{i}=\sum_{\mathbf{r}_{(\cdot)}}\dot{\rho}\left(\mathbf{r}_{(\cdot)}^{\mathrm{T}}\boldsymbol{\Sigma}_{(\cdot)}^{-1}\mathbf{r}_{(\cdot)}\right)\mathbf{J}_{\mathbf{r}_{(\cdot)}}^{\mathbf{x}_i\mathrm{T}}\boldsymbol{\Sigma}_{(\cdot)}^{-1}\mathbf{r}_{(\cdot)}
\end{aligned}
$$

where λ is a damping factor; a lower λ makes the LM method dominated by the GN method. **H** is the information matrix of the increments $\Delta\mathbf{x}^*$, **I** is the identity matrix, and **b** is the corresponding error term. This method can control the step size on non-linear surfaces. Here, $\mathbf{H}_{ij}$ gives the ith row, jth column block matrix element of **H**, and $\mathbf{b}_i$ gives the ith row block vector element of **b**. $\rho()$ is the Huber robust cost function, which can alleviate the effect of outliers [17], and $\dot{\rho}$ () means its first-order derivative, which is derived by the chain rule. $\mathbf{r}_{(\cdot)}$ is the residual error vector of $(\cdot)$ that quantifies the mismatch between a measured quantity and the estimated value, and $\sum_{(\cdot)}$ is the corresponding covariance matrix of this error. $\mathbf{x}_i$ means the ith state vector. Therefore, $\mathbf{J}_{\mathbf{r}_{(\cdot)}}^{\mathbf{x}_i}$ means the Jacobian of $\mathbf{r}_{(\cdot)}$ with reference to (w.r.t.) $\mathbf{x}_i$.

In addition to using a right disturbance (RD) model [17], we also use a left disturbance (LD) model:

$$
\begin{aligned}
&\mathrm{RD}{:}\tilde{\mathbf{R}}=\mathbf{R}\mathrm{Exp}(\boldsymbol{\varepsilon});\mathrm{LD}{:}\tilde{\mathbf{R}}=\mathrm{Exp}(\boldsymbol{\varepsilon})\mathbf{R};\boldsymbol{\varepsilon}\ll 1\\
&\mathbf{R}=\mathrm{Exp}(\boldsymbol{\phi})=\mathbf{I}+\frac{\sin(\theta)}{\theta}\boldsymbol{\phi}^{\wedge}+\frac{1-\cos(\theta)}{\theta^{2}}\left(\boldsymbol{\phi}^{\wedge}\right)^{2},\boldsymbol{\phi}=\theta\mathbf{a} \qquad (2)\\
&\boldsymbol{\phi}=\mathrm{Log}(\mathbf{R})=\left(\frac{\theta\left(\mathbf{R}-\mathbf{R}^{\mathrm{T}}\right)}{2\sin(\theta)}\right)^{\vee},\theta=\cos^{-1}\left(\frac{\mathrm{tr}(\mathbf{R})-1}{2}\right)
\end{aligned}
$$

where **R** is a rotation matrix, which belongs to $\mathrm{SO}(3)\doteq$

$\{\mathbf{R}\in\mathbb{R}^{3\times3}:\mathbf{R}^{\mathrm{T}}\mathbf{R}=\mathbf{I},\det(\mathbf{R})=1\}$ , and $\boldsymbol{\varepsilon}$ is a small $\mathbb{R}^3$ perturbation, which belongs to so(3) or the Lie algebra of SO(3). Exp($\boldsymbol{\phi}$) is the "vectorized" exponential map [17] (Rodrigues formula): $\mathbb{R}^3\rightarrow$SO(3); $\boldsymbol{\phi}\rightarrow\exp(\boldsymbol{\phi}^\wedge)$, where $\boldsymbol{\phi}$ is so(3), **a** is a unit vector and the hat operator denotes $[\mathrm{x\ y\ z}]^{\mathrm{T}\wedge}=[0\ {-z}\ y,\ z\ 0\ {-x},\ {-y}\ x\ 0]$. Exp($\boldsymbol{\phi}$) is a bijection if $\|\boldsymbol{\phi}\|<\pi$, and the corresponding inverse mapping is the "vectorized" logarithm map Log($\mathbf{R}$), SO(3)$\rightarrow\mathbb{R}^3$; $\mathbf{R}\rightarrow\log(\mathbf{R})^\vee$. For a skew symmetric matrix $S=\boldsymbol{\phi}^\wedge$, the vee operator is such that $S^\vee=\boldsymbol{\phi}$. If we do not restrict the domain, Exp($\boldsymbol{\phi}$) becomes surjective as $\boldsymbol{\phi}=(\theta+2k\pi)\mathbf{a}$, $k\in\mathbb{Z}$ [17]. The disturbance model is independently used in 3 parts: (i) for the rotation increment $\Delta\mathbf{R}$, we maintain the RD, and ε is normally distributed with a zero mean and covariance Σ [17]; (ii) for the update of small delta measurements $\delta\mathbf{b}^{g}_{i}$, we maintain the RD [17]; (iii) for retractions in the LM method, we use both models. The similarity is that for poses such as $\mathbf{T}_{\mathrm{WB}}$ or $\mathbf{T}_{\mathrm{CW}}$, we always disturb the non-world frame, i.e., RD for $\mathbf{T}_{\mathrm{WB}}$ and LD for $\mathbf{T}_{\mathrm{CW}}$. To simplify the notation of the discussion, in the following part, we only use the RD form.

In the IMU/encoder error term, we use the same retraction of a small vector in g2o for SO(3) as [17]. For SE(3), we use a simpler retraction at **T=(p, R)**, with δ**p** instead of **R**δ**p.**

$$\mathrm{R_T}\left(\delta\mathbf{p},\delta\boldsymbol{\phi}\right)=\left(\mathbf{p}+\delta\mathbf{p},\mathbf{R}\mathrm{Exp}\left(\delta\boldsymbol{\phi}\right)\right),\left[\delta\mathbf{p},\delta\boldsymbol{\phi}\right]\in\mathbb{R}^6 \tag{3}$$

where $\mathrm{R_T}$ is the retraction function of **T**, $\mathbb{R}^{4\times4}$ SE(3), which consists of translation $\mathbb{R}^3$ **p** and rotation SO(3) **R** or its so(3) $\boldsymbol{\phi}$; δ(·) denotes a small variable of (·).

The MAP state estimation is similar to [17]. We add only one encoder term $E_{ij}$ to the set of measurements up to time $k$.

$$\mathrm{Z}_k\doteq\left\{C_i,I_{ij},E_{ij}\right\}_{(i,j)\in\mathrm{K}_k} \tag{4}$$

where $I_{\mathrm{ij}}$ and $E_{ij}$ are the set of IMU and encoder measurements between two consecutive keyframes (or frames) $i$ and $j$, respectively. $C_i$ is the image measurement at keyframe $i$ and contains the multiple image measurements $\mathbf{z}_{il}$. $\mathrm{K}_k$ is the set of keyframes up to time $k$.

Therefore, the negative log-posterior is:

$$\begin{aligned}\mathrm{X}_k^*&=\arg\min_{\mathrm{X}_k}-\ln p\left(\mathrm{X}_k\,|\,\mathrm{Z}_k\right)=\arg\min_{\mathrm{X}_k}\left\|\mathbf{r}_0\right\|^2_{\Sigma_0}+\sum_{i\in\mathrm{K}_k}\sum_{l\in C_i}\left\|\mathbf{r}_{\mathbf{z}_{il}}\right\|^2_{\Sigma_{\mathbf{z}_{il}}}\\&+\sum_{(i,j)\in\mathrm{K}_k}(\left\|\mathbf{r}_{I_{ij}}\right\|^2_{\Sigma_{I_{ij}}}+\left\|\mathbf{r}_{E_{ij}}\right\|^2_{\Sigma_{E_{ij}}}),\left\|\mathbf{r}_{(\cdot)}\right\|^2_{\Sigma_{(\cdot)}}=\rho\left(\mathbf{r}_{(\cdot)}^{\mathrm{T}}\boldsymbol{\Sigma}_{(\cdot)}^{-1}\mathbf{r}_{(\cdot)}\right)\end{aligned} \tag{5}$$

where $\mathbf{X}_{\mathrm{k}}$ is the state set of all keyframes, which consists of states $\mathbf{x}_i$ of each keyframe $i$. Here, $\mathbf{z}_{il}$ indicates the image measurement of landmark $l$ at keyframe $i$. $\mathbf{r}_0$ is a priori residual error and is only used as a prior constraint $\mathrm{E_{prior}}$(j) of the state of the last frame j in our code, which uses an IMU as [16] does.

In VEOS2, we use the SE(3) of the camera to express the state:

$$\mathbf{x}_i=\mathbf{x}_{\mathrm{WC}i}\doteq\left[\mathbf{p}_i,\mathbf{R}_i\right]=\left[\mathbf{p}_{\mathrm{WC}i},\mathbf{R}_{\mathrm{WC}i}\right] \tag{6}$$

where $\mathbf{T}_{\mathrm{WC}i}$ is the pose of keyframe i, which maps a point from camera keyframe i to a word frame and consists of translation $\mathbf{p}_{\mathrm{WC}i}$ and SO(3) $\mathbf{R}_{\mathrm{WC}i}$. The RGB-D camera frame is noted as "C", and the world frame is noted as "W". Additionally, the left part denotes the IMU frame as "B" and the encoder frame (at the center of 2 wheels) as "E". The sensors are fixed, and their relative transform matrices $\mathbf{T}_{\mathrm{BC}}$ and $\mathbf{T}_{\mathrm{CE}}$ are known from prior calibration, such as a well-designed measurement or the Kalibr toolbox [20].

The VIEOS2 state definition of the system at time i is identical to that of VIOS2, which uses [17].

$$\mathbf{x}_i=\mathbf{x}_{\mathrm{WB}i}\doteq\left[\mathbf{p}_i,\mathbf{R}_i,\mathbf{v}_i,\mathbf{b}_i\right]=\left[\mathbf{p}_{\mathrm{WB}i},\mathbf{R}_{\mathrm{WB}i},\mathbf{v}_{\mathrm{WB}i},\mathbf{b}_i\right] \tag{7}$$

where SE(3) $\mathbf{T}_{\mathrm{WB}i}$ consists of translation $\mathbf{p}_{\mathrm{WB}i}$ and SO(3) $\mathbf{R}_{\mathrm{WB}i}$. Here, the state introduces the velocity $\mathbf{v}_{\mathrm{WB}i}$ of each keyframe i and the corresponding IMU bias $\mathbb{R}^6$ $\mathbf{b}_i=[\mathbf{b}^{g}_{i},\ \mathbf{b}^{a}_{i}]$, where the gyroscope bias is $\mathbf{b}^{g}_{i}\in\mathbb{R}^3$ and the accelerometer bias is $\mathbf{b}^{a}_{i}\in\mathbb{R}^3$.

Our image/IMU measurement error terms are similar to those in [16] and [17]. The reprojection residual error in (5) is defined as:

$$\begin{aligned}&\mathbf{r}_{\mathbf{z}_{il}}=\mathbf{z}_{il}-\pi\left({}_{\mathrm{C}i}\mathbf{p}_l\right)\\&{}_{\mathrm{C}i}\mathbf{p}_l=\left[x_{il}\quad y_{il}\quad z_{il}\right]^{\mathrm{T}}=\mathbf{R}_{\mathrm{CB}}\mathbf{R}_i^{\mathrm{T}}\left({}_{\mathrm{W}}\mathbf{p}_l-\mathbf{p}_i\right)+\mathbf{p}_{\mathrm{CB}}\\&\text{stereo/RGB-D:}\pi\left({}_{\mathrm{C}i}\mathbf{p}_l\right)=\left[u_{\mathrm{L}il}\quad v_{\mathrm{L}il}\quad u_{\mathrm{R}il}\right]^{\mathrm{T}}=\\&\left[f_x x_{il}/z_{il}+c_x\quad f_y x_{il}/z_{il}+c_y\quad u_{\mathrm{L}il}-bf_x/z_{il}\right]^{\mathrm{T}}\\&\mathbf{z}_{il}=\left[\tilde{u}_{\mathrm{L}il}\quad\tilde{v}_{\mathrm{L}il}\quad\tilde{u}_{\mathrm{L}il}-bf_x/\tilde{d}_{il}\right]^{\mathrm{T}}\\&\text{monocular:}\pi\left({}_{\mathrm{C}i}\mathbf{p}_l\right)=\left[u_{\mathrm{L}il}\quad v_{\mathrm{L}il}\right]^{\mathrm{T}},\mathbf{z}_{il}=\left[\tilde{u}_{\mathrm{L}il}\quad\tilde{v}_{\mathrm{L}il}\right]^{\mathrm{T}}\end{aligned} \tag{8}$$

where π() is the reprojection function, ${}_{\mathrm{C}i}\mathrm{p}_l$ is the $\mathbb{R}^3$ position of map point $l$ in the camera i frame, and ${}_{\mathrm{W}}\mathrm{p}_l$ is the position in the world frame. $f_x$, $f_y$, $c_x$, $c_y$ are camera intrinsic parameters, the focuses and center coordinates, respectively. $b$ is the horizontal shift between two rectified stereo cameras or the baseline between the infrared (IR) projector and the IR camera [9]. $\mathbf{z}_{il}$ is the $\mathbb{R}^2$ feature coordinate of landmark $l$ in keyframe $i$ for monocular cameras and $\mathbb{R}^3$ for stereo or RGB-D cameras, where one more term $\tilde{d}_{il}$ is the depth measurement. For VI(E)OS2, $\mathbf{R}_{\mathrm{CB}}$ and $\mathbf{p}_{\mathrm{CB}}$ are obtained from the inverse of $\mathbf{T}_{\mathrm{BC}}$; for VEOS2, they are obtained from the identity matrix $\mathrm{I}_{4\times4}$ because state $\mathbf{x}_i$ implies $\mathbf{x}_{\mathrm{WB}i}$ in VI(E)OS2 but $\mathbf{x}_{\mathrm{WC}i}$ in VEOS2.

We use an identical IMU measurement model to that in [17], where only the angle/velocity random walk noise ($\boldsymbol{\eta}^{g}$, $\boldsymbol{\eta}^{a}$) and rate random walk noise ($\boldsymbol{\eta}^{bg}$, $\boldsymbol{\eta}^{ba}$) are considered [21]:

$$\begin{aligned}&\tilde{\boldsymbol{\omega}}_{\mathrm{I}}(t)={}_{\mathrm{B}}\tilde{\boldsymbol{\omega}}_{\mathrm{WB}}(t)={}_{\mathrm{B}}\boldsymbol{\omega}_{\mathrm{WB}}(t)+\mathbf{b}^{\mathrm{g}}(t)+\boldsymbol{\eta}^{\mathrm{g}}(t)\\&\tilde{\mathbf{a}}_{\mathrm{I}}(t)={}_{\mathrm{B}}\tilde{\mathbf{a}}_{\mathrm{WB}}(t)={}_{\mathrm{B}}\mathbf{a}_{\mathrm{WB}}(t)+\mathbf{b}^{\mathrm{a}}(t)+\boldsymbol{\eta}^{\mathrm{a}}(t)\\&{}_{\mathrm{B}}\mathbf{a}_{\mathrm{WB}}(t)=\mathbf{R}_{\mathrm{WB}}^{\mathrm{T}}(t)\left({}_{\mathrm{W}}\mathbf{a}_{\mathrm{WB}}(t)-{}_{\mathrm{W}}\mathbf{g}\right)\\&\dot{\mathbf{b}}^{\mathrm{g}}(t)=\boldsymbol{\eta}^{\mathrm{bg}}(t),\dot{\mathbf{b}}^{\mathrm{a}}(t)=\boldsymbol{\eta}^{\mathrm{ba}}(t)\end{aligned} \tag{9}$$

$\tilde{\boldsymbol{\omega}}_{\mathrm{I}}(t)$ and $\tilde{\mathbf{a}}_{\mathrm{I}}(t)$ are the 3-axis gyroscope and accelerometer measurements of the IMU at time $t$, respectively. They are affected by the additive white noise $\boldsymbol{\eta}$ and a slowly varying sensor bias **b**. The vector ${}_{\mathrm{B}}\boldsymbol{\omega}_{\mathrm{WB}}(t)$ is the instantaneous angular velocity of B relative to W expressed in coordinate frame B, while ${}_{\mathrm{B}}\mathbf{a}_{\mathrm{WB}}(t)$ is the acceleration of the sensor; ${}_{\mathrm{W}}\mathbf{g}$ is the gravity vector in world coordinates. We also neglect the effects due to the Earth's rotation, so we assume that W is an inertial frame [17].

The IMU residual error in (5) is defined as:

$$\begin{aligned}&\mathbf{r}_{I_{ij}}=\left[\mathbf{r}_{Inb_{ij}}\quad\mathbf{r}_{b_{ij}}\right]^{\mathrm{T}},\mathbf{r}_{Inb_{ij}}=\left[\mathbf{e}_{\mathrm{Ip}}^{\mathrm{T}}\quad\mathbf{e}_{\mathrm{IR}}^{\mathrm{T}}\quad\mathbf{e}_{\mathrm{Iv}}^{\mathrm{T}}\right]^{\mathrm{T}},\mathbf{r}_{b_{ij}}=\left[\mathbf{b}_j-\mathbf{b}_i\right]^{\mathrm{T}}\\&\mathbf{e}_{\mathrm{Ip}}=\mathbf{R}_i^{\mathrm{T}}\left(\mathbf{p}_j-\mathbf{p}_i-\mathbf{v}_i\Delta t_{ij}-\frac{\mathbf{g}\Delta t_{ij}^2}{2}\right)-\left(\Delta\bar{\mathbf{p}}_{ij}+\mathbf{J}^{\mathrm{g}}_{\Delta\mathbf{p}_{ij}}\delta\mathbf{b}_i^{\mathrm{g}}+\mathbf{J}^{\mathrm{a}}_{\Delta\mathbf{p}_{ij}}\delta\mathbf{b}_i^{\mathrm{a}}\right)\\&\mathbf{e}_{\mathrm{IR}}=\mathrm{Log}\left(\left(\Delta\bar{\mathbf{R}}_{ij}\mathrm{Exp}\left(\mathbf{J}^{\mathrm{g}}_{\Delta\mathbf{R}_{ij}}\delta\mathbf{b}_i^{\mathrm{g}}\right)\right)^{\mathrm{T}}\mathbf{R}_i^{\mathrm{T}}\mathbf{R}_j\right)\end{aligned}$$

$$\mathbf{e}_{\mathrm{Iv}} = \mathbf{R}_i^{\mathrm{T}}\left(\mathbf{v}_j - \mathbf{v}_i - \mathbf{g}\Delta t_{ij}\right) - \left(\Delta\bar{\mathbf{v}}_{ij} + \mathbf{J}_{\Delta \mathbf{v}_{ij}}^{\mathrm{g}}\delta\mathbf{b}_i^{\mathrm{g}} + \mathbf{J}_{\Delta \mathbf{v}_{ij}}^{\mathrm{a}}\delta\mathbf{b}_i^{\mathrm{a}}\right) \quad (10)$$

where the residual error can be directly computed from state $\mathbf{x}_i$, $\mathbf{x}_j$ without time-consuming integration when the states are changed by defining these relative states, whose reference frame $B_i$ is a non-inertial frame with acceleration $\mathbf{g}$ and velocity $\mathbf{v}_i$. The bias term can be independent of the IMU without a bias term, which implies that $\left\|\mathbf{r}_{I_{ij}}\right\|_{\Sigma_{I_{ij}}}^2 = \left\|\mathbf{r}_{Inb_{ij}}\right\|_{\Sigma_{Inb_{ij}}}^2 + \left\|\mathbf{r}_{b_{ij}}\right\|_{\Sigma_{b_{ij}}}^2$, if a small change $\delta\mathbf{b}_i$ is introduced at linearization point $\mathbf{x}_i$ of the preintegration term, where $\mathbf{b}_i = \bar{\mathbf{b}}_i + \delta\mathbf{b}_i$, and using a 1st-order expansion [17]. The preintegration term $\Delta\bar{\mathbf{p}}_{ij} = \Delta\tilde{\mathbf{p}}_{ij}\left(\bar{\mathbf{b}}_i\right)$, $\Delta\bar{\mathbf{R}}_{ij} = \Delta\tilde{\mathbf{R}}_{ij}\left(\bar{\mathbf{b}}_i^{\mathrm{g}}\right)$, $\Delta\bar{\mathbf{v}}_{ij} = \Delta\tilde{\mathbf{v}}_{ij}\left(\bar{\mathbf{b}}_i\right)$, where $\bar{\mathbf{b}}_i = \left[\bar{\mathbf{b}}_i^{\mathrm{g}}, \bar{\mathbf{b}}_i^{\mathrm{a}}\right]$, and the linearization-related Jacobians $\mathbf{J}$ can be derived as follows [17]:

$$\Delta\bar{\mathbf{p}}_{ij} = \Delta\bar{\mathbf{p}}_{ij-1} + \Delta\bar{\mathbf{v}}_{ij-1}\Delta t + \Delta\bar{\mathbf{R}}_{ij-1}\hat{\mathbf{a}}_{\mathrm{I}j\text{-}1}\Delta t^2/2, \hat{\mathbf{a}}_{\mathrm{I}j\text{-}1} = \tilde{\mathbf{a}}_{\mathrm{I}j\text{-}1} - \bar{\mathbf{b}}_i^{\mathrm{a}} \quad (11)$$

$$\Delta\bar{\mathbf{v}}_{ij} = \Delta\bar{\mathbf{v}}_{ij-1} + \Delta\bar{\mathbf{R}}_{ij-1}\hat{\mathbf{a}}_{\mathrm{I}j\text{-}1}\Delta t, \Delta\bar{\mathbf{R}}_{ij} = \Delta\bar{\mathbf{R}}_{ij-1}\mathrm{Exp}\left(\hat{\boldsymbol{\omega}}_{\mathrm{I}j\text{-}1}\Delta t\right)$$

$$\hat{\boldsymbol{\omega}}_{\mathrm{I}j\text{-}1} = \tilde{\boldsymbol{\omega}}_{\mathrm{I}j\text{-}1} - \bar{\mathbf{b}}_i^{\mathrm{g}}; \frac{\partial\Delta\overline{(\cdot)}_{ij}}{\partial\mathbf{b}_i^{(\cdot)}} = \frac{\partial\Delta\widetilde{(\cdot)}_{ij}\left(\mathbf{b}_i^{(\cdot)}\right)}{\partial\mathbf{b}_i^{(\cdot)}}\left(\bar{\mathbf{b}}_i^{(\cdot)}\right)$$

$$\mathbf{J}_{\Delta\mathbf{p}_{ij}}^{\mathrm{g}} = \frac{\partial\Delta\bar{\mathbf{p}}_{ij}}{\partial\mathbf{b}_i^{\mathrm{g}}} = \mathbf{J}_{\Delta\mathbf{p}_{ij-1}}^{\mathrm{g}} + \mathbf{J}_{\Delta\mathbf{v}_{ij-1}}^{\mathrm{g}}\Delta t - \Delta\bar{\mathbf{R}}_{ij-1}\hat{\mathbf{a}}_{\mathrm{I}j\text{-}1}{}^{\wedge}\mathbf{J}_{\Delta\mathbf{R}_{ij-1}}^{\mathrm{g}}\Delta t^2/2$$

$$\mathbf{J}_{\Delta\mathbf{p}_{ij}}^{\mathrm{a}} = \frac{\partial\Delta\bar{\mathbf{p}}_{ij}}{\partial\mathbf{b}_i^{\mathrm{a}}} = \mathbf{J}_{\Delta\mathbf{p}_{ij-1}}^{\mathrm{a}} + \mathbf{J}_{\Delta\mathbf{v}_{ij-1}}^{\mathrm{a}}\Delta t - \Delta\bar{\mathbf{R}}_{ij-1}\Delta t^2/2$$

$$\mathbf{J}_{\Delta\mathbf{v}_{ij}}^{\mathrm{g}} = \frac{\partial\Delta\bar{\mathbf{v}}_{ij}}{\partial\mathbf{b}_i^{\mathrm{g}}} = \mathbf{J}_{\Delta\mathbf{v}_{ij-1}}^{\mathrm{g}} - \Delta\bar{\mathbf{R}}_{ij-1}\hat{\mathbf{a}}_{\mathrm{I}j\text{-}1}{}^{\wedge}\mathbf{J}_{\Delta\mathbf{R}_{ij-1}}^{\mathrm{g}}\Delta t$$

$$\mathbf{J}_{\Delta\mathbf{v}_{ij}}^{\mathrm{a}} = \frac{\partial\Delta\bar{\mathbf{v}}_{ij}}{\partial\mathbf{b}_i^{\mathrm{a}}} = \mathbf{J}_{\Delta\mathbf{v}_{ij-1}}^{\mathrm{a}} - \Delta\bar{\mathbf{R}}_{ij-1}\Delta t$$

$$\mathbf{J}_{\Delta\mathbf{R}_{ij}}^{\mathrm{g}} = \frac{\partial\Delta\bar{\mathbf{R}}_{ij}}{\partial\mathbf{b}_i^{\mathrm{g}}} = \Delta\bar{\mathbf{R}}_{j-1j}^{\mathrm{T}}\mathbf{J}_{\Delta\mathbf{R}_{ij-1}}^{\mathrm{g}} - \mathrm{J}_{\mathrm{r}}^{j-1}\Delta t$$

where $(\cdot)^{(\cdot)}{}_{(\cdot)j\text{-}1} = (\cdot)^{(\cdot)}{}_{(\cdot)}(t_{j\text{-}1})$; strictly speaking, $\Delta t$ should be $\Delta t_{j\text{-}1j}$. $\mathrm{J}_{\mathrm{r}}^{j-1} = \mathrm{J}_{\mathrm{r}}\left(\hat{\boldsymbol{\omega}}_{\mathrm{I}j\text{-}1}\Delta t\right)$, where $\mathrm{J}_{\mathrm{r}}()$ is the right Jacobian of SO(3) [17] and $\Delta\bar{\mathbf{R}}_{j-1j} = \mathrm{Exp}\left(\hat{\boldsymbol{\omega}}_{\mathrm{I}j\text{-}1}\Delta t\right)$. We provide a clear iterative form that is helpful for coding, while [17] did not. The iterative noise propagation and analytic Jacobians in the optimization are provided in the appendix of [17]. We only multiply $\mathbf{R}^{\mathrm{T}}_{(\cdot)}$ by the right side of each Jacobian w.r.t. $\delta\mathbf{p}_{(\cdot)}$ since our retraction of $\mathbf{p}$ is different (3).

## III. VIEOS2 FLOW CHART

The basic procedure and our IMU initialization method of visual-inertial SLAM are similar to those of the ORBSLAM2 code [9] and VIORBSLAM paper [16]; only the new part of our flow chart is shown in Fig. 1.

Two parallel threads are added: caching encoder data and IMU data. The nonlinear optimization functions are extended for the encoder: motion-only BA, local BA, full BA, and pose graph optimization. The motion-only BA occurs in the pose prediction and local tracking map. One new requirement is to synchronize the encoders or IMU data with frames by using a list and a certain permitted aligning time error. The IMU data between 2 keyframes will be discarded when the nearest time of IMU data to keyframe i is over this error.

One important change is to enable the addition of pure encoder edges, which connect 2 states $\mathbf{x}_i$ and $\mathbf{x}_j$ by only encoder measurements in VEOS2 but encoder and IMU measurements in VIEOS2, to the spanning tree, which is mainly used to propagate the corrected poses to all keyframes in the map. The covariance of pure encoder edges in the pose graph, which requires a fast process and only optimizes all states $\mathbf{x}_i$ by measuring the loops and relative states, is set by a simple but effective strategy. The minimum covariance among pure encoder edges will be found. Assuming that it can reach the same accuracy of another visual edge, which at least uses image measurements, its relative covariance is set on each pure encoder edge. The pure encoder edges with a larger covariance are expected to be changed more to reduce their large error. We also prevent consecutive pure encoder edges from introducing too many inaccurate map points. Finally, we lower the threshold of loop detection after introducing pure encoder edges but change it back when loop closure occurs. This type of edge also helps to establish a robust IMU initialization in VIEOS2 despite visual failure.

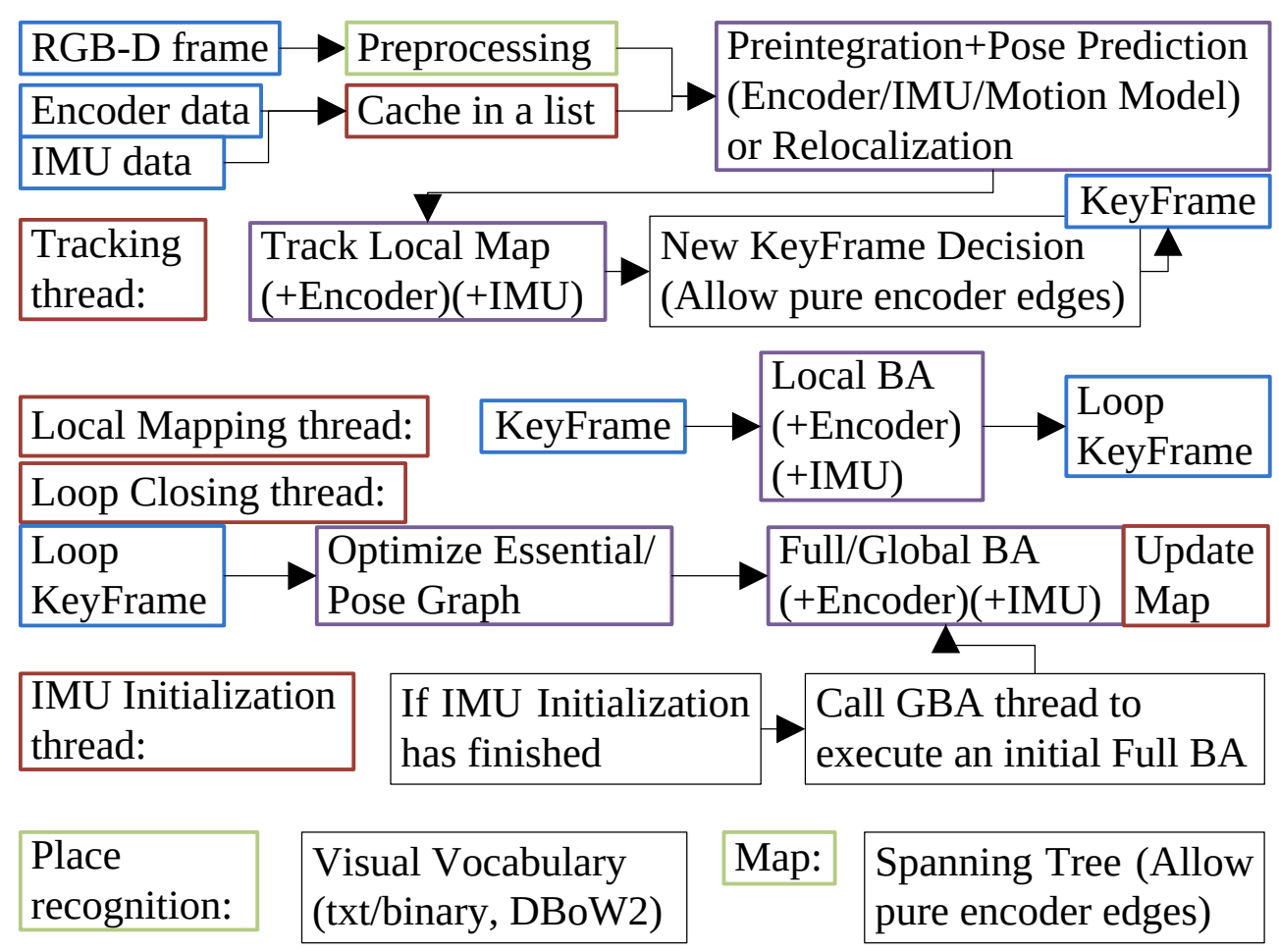

Fig. 1. Flow chart of VIEOS2.

We now explain the extended part of our BA, as shown in Fig. 2. $\mathbf{P}$ is the pose ($\mathbf{p}_i$, $\mathbf{R}_i$), $\mathbf{v}$ is the velocity, and $\mathbf{b}$ is the IMU biases. Orange indicates that the state is fixed or out of total states $\mathbf{X}_k$, yellow indicates that the state is optimized or in $\mathbf{X}_k$, and the current keyframe in the local BA is denoted as the last 0th keyframe, which indicates that the last Nth keyframe is over the sliding window of N last keyframes, and its state must be fixed. The states of the map points are ${}_{\mathrm{w}}\mathbf{p}_l$. The motion-only BA only optimizes the states of keyframes $\mathbf{x}_i$, but the local BA also optimizes the states of map points ${}_{\mathrm{w}}\mathbf{p}_l$, and the Schur complement is used in g2o to speed up the process [19]. We do not add an encoder error between $\mathbf{v}$ for 2 reasons. First, $\boldsymbol{\omega}_i$ must be added to state $\mathbf{x}_i$ to transform $\mathbf{v}$ from the IMU to the encoder, which costs more time. Second, the product of the allowed maximum $\Delta t_{ij}$ between 2 keyframes and $\Delta t$ between 2 IMU measurements is small (3/200 in our case), so the covariance of the encoder error between $\mathbf{v}$ is much smaller than the integrated error between $\mathbf{P}$, which indicates that the encoder error between $\mathbf{v}$ can be ignored.

We only introduced the motion-only BA in VEOS2 and the local BA in VIEOS2, since the local BA in VEOS2 simply omits the IMU error terms, and the motion-only BA in VIEOS2 is similar to the optimization of Fig. 2 [16], not including the added encoder error terms and using a prior error term $\mathbf{r}_0$ in (5).

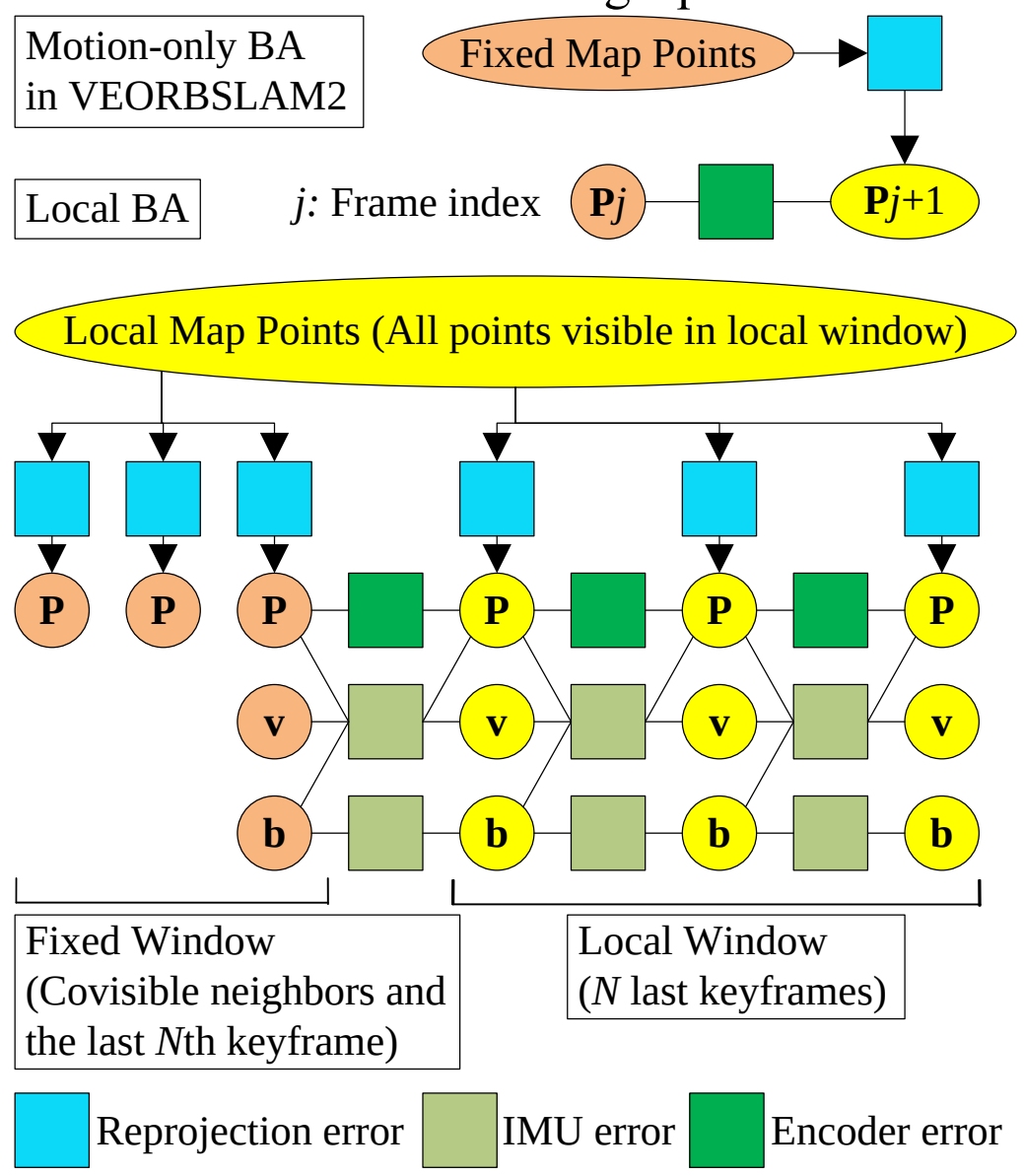

Fig. 2. Motion-only BA of VEOS2 and local BA of VIEOS2.

We also constrain the interval of two consecutive keyframes in the local window of the local BA to be less than 0.5 s and any two consecutive keyframes to be less than 3 s in VI(E)OS2. The full BA optimizes the states of all keyframes and map points in the map and introduces one more common variable named scale to obtain a better result, especially in the EuRoC dataset.

## IV. Encoder Preintegration

### A. Measurement Model and Preintegration

These measurement assumptions are the theoretical basis. Let the forward direction of the two velocities point to the front or the +x-axis of the encoder frame "E". We denote ${}_{\mathrm{E}}(\cdot)_{\mathrm{WE}}(t)$ or ${}_{\mathrm{E}(t)}(\cdot)_{\mathrm{E}(t)\mathrm{E}(t+dt)}(t)$ as $(\cdot)_{\mathrm{EE}}(t)$, where the latter E($t$+d$t$) denotes the coordinate system fixed on the encoder frame and the former E($t$) is the encoder frame, which is still relative to W at time $t$.

$$\tilde{v}^{\mathrm{l}}_{\mathrm{EE}}(t)=v^{\mathrm{l}}_{\mathrm{EE}}(t)+\eta^{\mathrm{el}}(t),\tilde{v}^{\mathrm{r}}_{\mathrm{EE}}(t)=v^{\mathrm{r}}_{\mathrm{EE}}(t)+\eta^{\mathrm{er}}(t) \tag{12}$$

where $[\eta^{\mathrm{el}}\ \eta^{\mathrm{er}}]^{\mathrm{T}}\sim \mathrm{N}(\mathbf{0}, \Sigma_{\mathrm{e}}\in\mathbb{R}^{2\times2})$ is the additive white noise (zero mean and covariance $\Sigma_{\mathrm{e}}$) caused by the sensor measurement. $\mathbf{v}^{\mathrm{l}}_{\mathrm{EE}}(t)$ and $\mathbf{v}^{\mathrm{r}}_{\mathrm{EE}}(t)$ are the instantaneous velocity estimates of the left and right wheel centers of E relative to W expressed in coordinate frame E at time $t$, respectively. $\tilde{v}^{(\cdot)}_{\mathrm{EE}}$ is the corresponding encoder measurement.

We can use a right-handed coordinate system and kinematic model of a differential wheeled robot under the conditions that the motion between two consecutive keyframes is limited to an almost flat plane and there is no sliding friction during the motion.

$$\dot{\mathbf{R}}_{\mathrm{E}i\mathrm{E}}=\mathbf{R}_{\mathrm{E}i\mathrm{E}}\boldsymbol{\omega}_{\mathrm{EE}}{}^{\wedge},\dot{\mathbf{p}}_{\mathrm{E}i\mathrm{E}}=\mathbf{v}_{\mathrm{E}i\mathrm{E}} \tag{13}$$

where $(\cdot)_{\mathrm{E}i\mathrm{E}}$ denotes ${}_{\mathrm{E}i}(\cdot)_{\mathrm{E}i\mathrm{E}}$ and $\boldsymbol{\omega}_{\mathrm{EE}}$ denotes ${}_{\mathrm{E}}\boldsymbol{\omega}_{\mathrm{E}i\mathrm{E}}(t)$. E$i$ is the encoder frame at time $t_i$, which is still relative to W/E($t$).

As Fig. 3 shows, the forward velocity $\mathbf{v}_{\mathrm{EE}}=[v\ 0\ 0]^{\mathrm{T}}-\boldsymbol{\eta}^{\mathrm{mv}}$, where $\mathbb{R}^3\ \boldsymbol{\eta}^{\mathrm{mv}}=[\eta^{\mathrm{mvx}}\ \eta^{\mathrm{mvy}}\ \eta^{\mathrm{mvz}}]^{\mathrm{T}}$, and the rotational velocity $\boldsymbol{\omega}_{\mathrm{EE}}=[0\ 0\ \omega]^{\mathrm{T}}-\boldsymbol{\eta}^{\mathrm{m\omega}}$, where $\mathbb{R}^3\ \boldsymbol{\eta}^{\mathrm{m\omega}}=[\eta^{\mathrm{m\omega x}}\ \eta^{\mathrm{m\omega y}}\ \eta^{\mathrm{m\omega z}}]^{\mathrm{T}}$, can be derived from the speed measurements of two wheels. Here, the noise $\mathbb{R}^6\ \boldsymbol{\eta}^{\mathrm{m}}=[\boldsymbol{\eta}^{\mathrm{mv}},\ \boldsymbol{\eta}^{\mathrm{m\omega}}]$ is caused by the error of the kinematic model, the prior calibration of $\mathbf{T}_{\mathrm{CE}}$, and the flat plane and friction assumptions. Therefore, we assume that these encoder model noises are normally distributed perturbations with a zero mean and covariance $\Sigma_{\mathrm{em}}\in\mathbb{R}^{6\times6}$.

$$\begin{bmatrix}\mathbf{v}(t)\\ \boldsymbol{\omega}(t)\end{bmatrix}=\frac{1}{2}\begin{bmatrix}\left(v^{\mathrm{l}}_{\mathrm{EE}}(t)+v^{\mathrm{r}}_{\mathrm{EE}}(t)\right)\mathbf{i}\\ \frac{1}{r_c}\left(v^{\mathrm{r}}_{\mathrm{EE}}(t)-v^{\mathrm{l}}_{\mathrm{EE}}(t)\right)\mathbf{i}\times\mathbf{j}\end{bmatrix}=\begin{bmatrix}v(t)\mathbf{i}\\ \omega(t)\mathbf{k}\end{bmatrix}$$
$$v(t)=\frac{v^{\mathrm{l}}_{\mathrm{EE}}(t)+v^{\mathrm{r}}_{\mathrm{EE}}(t)}{2},\omega(t)=\frac{v^{\mathrm{r}}_{\mathrm{EE}}(t)-v^{\mathrm{l}}_{\mathrm{EE}}(t)}{2r_c} \tag{14}$$

where $r_c$ is half the distance between two wheels and $\mathbf{j}$ is the y-axis unit vector of the encoder frame "E". The forward directions of $\mathbf{v}$ and $\boldsymbol{\omega}$ point to the front and +z-axis of frame E, respectively. $\mathbf{v}$ is the average of the velocities of the 2 wheels, and $\omega r_c$ is the tangent speed or half-speed difference of the 2 wheels.

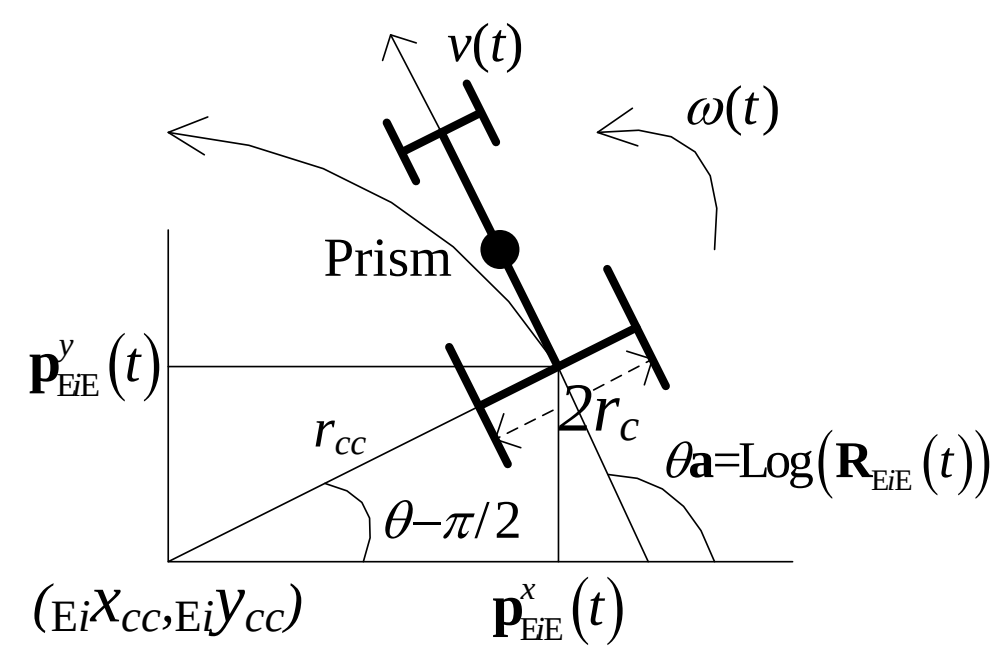

Fig. 3. Kinematic model of a differential wheeled robot. ${}_{\mathrm{E}i}x_{\mathrm{CC}}$ and ${}_{\mathrm{E}i}y_{\mathrm{CC}}$ are the rotation circle center coordinates in frame E$i$, and $r_{\mathrm{CC}}$ is the corresponding rotation radius.

By integrating the kinematic model (13), we obtain the state at time $t+\Delta t$.

$$\mathbf{R}_{\mathrm{E}i\mathrm{E}}(t+\Delta t)=\mathbf{R}_{\mathrm{E}i\mathrm{E}}(t)\mathrm{Exp}\left(\boldsymbol{\omega}_{\mathrm{EE}}(t)\Delta t\right)$$
$$\mathbf{p}_{\mathrm{E}i\mathrm{E}}(t+\Delta t)=\mathbf{p}_{\mathrm{E}i\mathrm{E}}(t)+\int_{t}^{t+\Delta t}\mathbf{R}_{\mathrm{E}i\mathrm{E}}(s)\mathbf{v}_{\mathrm{EE}}(s)ds \tag{15}$$

where $\boldsymbol{\omega}_{\mathrm{EE}}(t)\Delta t=(\boldsymbol{\omega}-\boldsymbol{\eta}^{\mathrm{m\omega d}})\Delta t=-[\eta^{\mathrm{m\omega dx}}(t)\ \eta^{\mathrm{m\omega dy}}(t)\ \eta^{\mathrm{m\omega dz}}(t)-\omega(t)]^{\mathrm{T}}\Delta t$. $(\cdot)^{(\cdot)\mathrm{d}}$ denotes discrete-time noise, whose covariance is different from that of continuous-time noise $(\cdot)^{(\cdot)}$, which will be discussed later. $(\cdot)^{(\cdot)\mathrm{x}}$, $(\cdot)^{(\cdot)\mathrm{y}}$, and $(\cdot)^{(\cdot)\mathrm{z}}$ denote the x, y, and z components of the $\mathbb{R}^3$ vector $(\cdot)^{(\cdot)}$, respectively.

Using (12) and (14), we can use encoder measurements to express $\boldsymbol{\omega}$ and $\mathbf{v}$; hence, we obtain (we skip the similar processes in [17] and use the 1st-order approximation and Rodrigues' formula [18]):

$$\mathbf{R}_{\mathrm{E}i\mathrm{E}}(t+\Delta t)=\mathbf{R}_{\mathrm{E}i\mathrm{E}}(t)\mathrm{Exp}\left(\left(\tilde{\boldsymbol{\omega}}_t-\boldsymbol{\eta}^{\omega\mathrm{d}}_t\right)\Delta t\right) \tag{16}$$
$$\approx\mathbf{R}_{\mathrm{E}i\mathrm{E}}(t)\mathrm{Exp}(\tilde{\boldsymbol{\omega}}_t\Delta t)\mathrm{Exp}\left(-\boldsymbol{\varepsilon}^{\mathrm{R}}_{t,\Delta t}\right),\boldsymbol{\varepsilon}^{\mathrm{R}}_{t,\Delta t}=\Delta t\mathrm{J}_{\mathrm{r}}(\tilde{\boldsymbol{\omega}}_t\Delta t)\boldsymbol{\eta}^{\omega\mathrm{d}}_t$$

where $(\cdot)^{(\cdot)}_t=(\cdot)^{(\cdot)}(t)$; $\boldsymbol{\varepsilon}^{\mathrm{R}}_{t,\Delta t}=\boldsymbol{\varepsilon}^{\mathrm{R}}(t,\ \Delta t)$ is a small rotational disturbance caused by noise during time interval ($t$, $t+\Delta t$]; $\tilde{\boldsymbol{\omega}}_t\Delta t=[0,0,\ \tilde{\omega}_t\Delta t\ ]$, $\tilde{\omega}_t=(\ \tilde{v}^{\mathrm{r}}_{\mathrm{EE}}(t)\ -\ \tilde{v}^{\mathrm{l}}_{\mathrm{EE}}(t)\ )/(\ 2r_c\ )$ is an so(3) whose rotational axis is z due to the almost-flat plane assumption during $\Delta t$, and its SO(3) is simple: Exp([0 0

$\theta]^{\mathrm{T}}$)=[cos$\theta$ –sin$\theta$ 0, sin$\theta$ cos$\theta$ 0, 0 0 1]. Its $\mathrm{J_r}$ is also simple: $\mathrm{J_r}$([0 0 $\theta]^{\mathrm{T}}$)=[sin$\theta/\theta$ (1-cos$\theta)/\theta$ 0, (cos$\theta$-1)/$\theta$ sin$\theta/\theta$ 0, 0 0 1]. $\boldsymbol{\eta}^{\omega d}=\boldsymbol{\eta}^{e\omega d}+\boldsymbol{\eta}^{m\omega d}$ is the total rotational noise caused by model noise $\boldsymbol{\eta}^{m\omega d}$ and encoder noise $\boldsymbol{\eta}^{e\omega d}=[0\ 0\ \eta^{e\omega d}]^{\mathrm{T}}$, where $\eta^{e\omega d}=(\eta^{erd}-\eta^{eld})/(2r_c)$. Here, $\mathrm{v_{EE}}^{l}(t)$, $\mathrm{v_{EE}}^{r}(t)$ and $\eta^{(\cdot)}(t)$ can remain constant or linear during time interval $\Delta t$; we only show the constant result and use the linear result in our code. An exact solution of (13) for measurements with a nonzero rotation rate due to the almost flat plane assumption instead of the approximate solution [17] is provided.

The covariance of the discrete-time encoder noise $[\eta^{eld}\ \eta^{erd}]^{\mathrm{T}}$ is also a function of the sampling rate and relates to the continuous-time spectral noise $[\eta^{el}\ \eta^{er}]^{\mathrm{T}}$ via $\sum_{ed}=\sum_{e}/\Delta t$. While the noise $\boldsymbol{\eta}^{m\omega d}$ caused by the model, calibration, or assumption should be positively correlated with the span between encoder measurements, intuitively, $\sum_{emd}=\Delta t\sum_{em}$; we can consider this type of error "Brownian motion" because it is assumed to have a slowly time-varying quantity [17].

$$\mathbf{R}_{\mathrm{E}i\mathrm{E}t}^{\mathrm{T}}\left(\mathbf{p}_{\mathrm{EiE}}\left(t+\Delta t\right)-\mathbf{p}_{\mathrm{EiE}}\left(t\right)\right)=\mathbf{R}_{\mathrm{EiE}}^{\mathrm{T}}\left(t\right)\int_{t}^{t+\Delta t}\mathbf{R}_{\mathrm{EiE}}\left(s\right)\left(\tilde{\mathbf{v}}_s-\boldsymbol{\eta}_s^{\mathrm{v}}\right)ds$$
$$\approx\int_0^{\Delta t}\mathrm{Exp}\left(\tilde{\boldsymbol{\omega}}_t s\right)\mathrm{Exp}\left(-\boldsymbol{\varepsilon}_{t,s}^{\mathrm{R}}\right)ds\left(\tilde{\mathbf{v}}_t-\boldsymbol{\eta}_t^{\mathrm{vd}}\right),\mathrm{Exp}\left(-\boldsymbol{\varepsilon}_{t,s}^{\mathrm{R}}\right)\approx\mathbf{I}-\boldsymbol{\varepsilon}_{t,s}^{\mathrm{R}\wedge}$$
(17)

where $(\cdot)^{(\cdot)}_s=(\cdot)^{(\cdot)}(s)$, $\tilde{v}_t=(\tilde{v}_{\mathrm{EE}}^{l}(t)+\tilde{v}_{\mathrm{EE}}^{r}(t))/2$, $\boldsymbol{\eta}^{vd}=\boldsymbol{\eta}^{evd}+\boldsymbol{\eta}^{mvd}$ is the total translation noise caused by the model noise $\boldsymbol{\eta}^{mvd}$ and encoder noise $\boldsymbol{\eta}^{evd}=[\eta^{evd}\ 0\ 0]^{\mathrm{T}}$; $\eta^{evd}=(\eta^{eld}+\eta^{erd})/2$. Only the x component of $\mathbf{v}_{\mathrm{EE}}$ is not small, and $\boldsymbol{\varepsilon}^{\mathrm{R}}_{t,s}$ is small; for the 1st-order approximation, only the 0th column of $\boldsymbol{\varepsilon}^{\mathrm{R}}_{t,s}{}^{\wedge}$ remains:

$$\approx\int_0^{\Delta t}\mathrm{Exp}\left(\tilde{\boldsymbol{\omega}}_t s\right)\left(\mathbf{I}+\begin{bmatrix}0 & -s\eta_t^{\omega dz} & \varepsilon_{t,s}^{\mathrm{Ry}}\\ & \mathbf{0}_{2\times3} & \end{bmatrix}^{\mathrm{T}}\right)ds\left(\tilde{\mathbf{v}}_t-\boldsymbol{\eta}_t^{\mathrm{vd}}\right) \quad (18)$$
$$=\int_0^{\Delta t}\begin{bmatrix}\cos\left(\tilde{\omega}_t s\right)+s\eta_t^{\omega dz}\sin\left(\tilde{\omega}_t s\right) & -\sin\left(\tilde{\omega}_t s\right) & 0\\ \sin\left(\tilde{\omega}_t s\right)-s\eta_t^{\omega dz}\cos\left(\tilde{\omega}_t s\right) & \cos\left(\tilde{\omega}_t s\right) & 0\\ \eta_t^{\omega dy}\mathrm{S}_{\tilde{\omega}_t,s}-\eta_t^{\omega dx}\mathrm{C}_{\tilde{\omega}_t,s} & 0 & 1\end{bmatrix}ds\left(\tilde{\mathbf{v}}_t-\boldsymbol{\eta}_t^{\mathrm{vd}}\right)$$
$$=\begin{bmatrix}\mathrm{S}_{\tilde{\omega}_t,\Delta t}+\frac{\eta_t^{\omega dz}}{\tilde{\omega}_t}\left(\mathrm{S}_{\tilde{\omega}_t,\Delta t}-\Delta t\cos\left(\tilde{\omega}_t\Delta t\right)\right) & -\mathrm{C}_{\tilde{\omega}_t,\Delta t} & 0\\ \mathrm{C}_{\tilde{\omega}_t,\Delta t}+\frac{\eta_t^{\omega dz}}{\tilde{\omega}_t}\left(\mathrm{C}_{\tilde{\omega}_t,\Delta t}-\Delta t\sin\left(\tilde{\omega}_t\Delta t\right)\right) & \mathrm{S}_{\tilde{\omega}_t,\Delta t} & 0\\ \frac{\eta_t^{\omega dy}}{\tilde{\omega}_t}\mathrm{C}_{\tilde{\omega}_t,\Delta t}-\frac{\eta_t^{\omega dx}}{\tilde{\omega}_t}\left(\Delta t-\mathrm{S}_{\tilde{\omega}_t,\Delta t}\right) & 0 & 1\end{bmatrix}\left(\tilde{\mathbf{v}}_t-\boldsymbol{\eta}_t^{\mathrm{vd}}\right)$$

where we use $\mathrm{S}_{\omega,s}=\sin(\omega s)/\omega$, $\mathrm{C}_{\omega,s}=(1-\cos(\omega s))/\omega$ for a compact expression. Then, we apply the 1st-order approximation and obtain:

$$\approx\tilde{v}_t\left[\mathrm{S}_{\tilde{\omega}_t,\Delta t}\quad \mathrm{C}_{\tilde{\omega}_t,\Delta t}\quad 0\right]^{\mathrm{T}}-\boldsymbol{\varepsilon}_{t,\Delta t}^{\mathrm{p}}, \quad (19)$$
$$\boldsymbol{\varepsilon}_{t,\Delta t}^{\mathrm{p}}=\begin{bmatrix}\eta_t^{\mathrm{vdx}}\mathrm{S}_{\tilde{\omega}_t,\Delta t}-\eta_t^{\mathrm{vdy}}\mathrm{C}_{\tilde{\omega}_t,\Delta t}\\ \eta_t^{\mathrm{vdx}}\mathrm{C}_{\tilde{\omega}_t,\Delta t}+\eta_t^{\mathrm{vdy}}\mathrm{S}_{\tilde{\omega}_t,\Delta t}\\ \eta_t^{\mathrm{vdz}}\Delta t\end{bmatrix}-\frac{\tilde{v}_t}{\tilde{\omega}_t}\begin{bmatrix}\eta_t^{\omega dz}\left(\mathrm{S}_{\tilde{\omega}_t,\Delta t}-\Delta t\cos\left(\tilde{\omega}_t\Delta t\right)\right)\\ \eta_t^{\omega dz}\left(\mathrm{C}_{\tilde{\omega}_t,\Delta t}-\Delta t\sin\left(\tilde{\omega}_t\Delta t\right)\right)\\ \eta_t^{\omega dx}\left(\mathrm{S}_{\tilde{\omega}_t,\Delta t}-\Delta t\right)+\eta_t^{\omega dy}\mathrm{C}_{\tilde{\omega}_t,\Delta t}\end{bmatrix}$$

where $\boldsymbol{\eta}^{vd}=[\eta^{vdx}\ \eta^{vdy}\ \eta^{vdz}]^{\mathrm{T}}=[\eta^{evd}+\eta^{mvdx}\ \eta^{mvdy}\ \eta^{mvdz}]^{\mathrm{T}}$ and $\boldsymbol{\eta}^{\omega d}=[\eta^{\omega dx}\ \eta^{\omega dy}\ \eta^{\omega dz}]^{\mathrm{T}}=[\eta^{m\omega dx}\ \eta^{m\omega dy}\ \eta^{e\omega d}+\eta^{m\omega dz}]^{\mathrm{T}}$. $\boldsymbol{\varepsilon}^{\mathrm{p}}_{t,\Delta t}=\boldsymbol{\varepsilon}^{\mathrm{p}}(t,\Delta t)$ is a small translation disturbance caused by noise during $(t, t+\Delta t]$.

We assume that the encoder is also synchronized with the camera and provides measurements at discrete times $k$ [17]. By iterating (16) and (19) for all $\Delta t$ intervals between two consecutive keyframes or frames at times $k=i$ and $k=j$, we can find the encoder preintegration.

$$\mathbf{R}_{\mathrm{E}i\mathrm{E}j}\approx\prod_{k=i}^{j\text{-}1}\mathrm{Exp}\left(\tilde{\boldsymbol{\omega}}_k\Delta t\right)\mathrm{Exp}\left(-\boldsymbol{\varepsilon}_{k,\Delta t}^{\mathrm{R}}\right)$$
$$\mathbf{p}_{\mathrm{E}i\mathrm{E}j}\approx\sum_{k=i}^{j\text{-}1}\mathbf{R}_{\mathrm{E}i\mathrm{E}k}\left(\tilde{v}_k\left[\mathrm{S}_{\tilde{\omega}_k,\Delta t}\quad \mathrm{C}_{\tilde{\omega}_k,\Delta t}\quad 0\right]^{\mathrm{T}}-\boldsymbol{\varepsilon}_{k,\Delta t}^{\mathrm{p}}\right) \quad (20)$$

where $(\cdot)^{(\cdot)}_{(\cdot)k}=(\cdot)^{(\cdot)}_{(\cdot)}(t_k)$, $\boldsymbol{\varepsilon}^{(\cdot)}_{k,\Delta t}=\boldsymbol{\varepsilon}^{(\cdot)}(t_k,\Delta t)$ and, strictly speaking, $\Delta t$ should be $\Delta t_{kk+1}$.

It is easy to define the following relative motion increments that are independent of linearization points $\mathbf{x}_i$ and $\mathbf{x}_j$, which is also the true relative physical change in pose, whose reference frame E$i$ is still relative to W.

$$\Delta\mathbf{R}_{\mathrm{E}i\mathrm{E}j}\doteq\mathbf{R}_{\mathrm{E}i\mathrm{E}i}^{\mathrm{T}}\mathbf{R}_{\mathrm{E}i\mathrm{E}j}=\mathbf{R}_{\mathrm{E}i\mathrm{E}j}$$
$$\Delta\mathbf{p}_{\mathrm{E}i\mathrm{E}j}\doteq\mathbf{R}_{\mathrm{E}i\mathrm{E}i}^{\mathrm{T}}\left(\mathbf{p}_{\mathrm{E}i\mathrm{E}j}-\mathbf{p}_{\mathrm{E}i\mathrm{E}i}\right)=\mathbf{p}_{\mathrm{E}i\mathrm{E}j} \quad (21)$$

Then, we must isolate the noise terms in (21) as [17] did.

$$\Delta\mathbf{R}_{\mathrm{E}i\mathrm{E}j}\approx\Delta\tilde{\mathbf{R}}_{\mathrm{E}i\mathrm{E}j}\prod_{k=i}^{j\text{-}1}\mathrm{Exp}\left(-\Delta\tilde{\mathbf{R}}_{\mathrm{E}k+1\mathrm{E}j}^{\mathrm{T}}\boldsymbol{\varepsilon}_{k,\Delta t}^{\mathrm{R}}\right)\doteq\Delta\tilde{\mathbf{R}}_{\mathrm{E}i\mathrm{E}j}\mathrm{Exp}\left(-\delta\boldsymbol{\phi}_{\mathrm{E}i\mathrm{E}j}\right)$$
$$\Delta\mathbf{p}_{\mathrm{E}i\mathrm{E}j}\approx\Delta\tilde{\mathbf{p}}_{\mathrm{E}i\mathrm{E}j}+\sum_{k=i}^{j\text{-}1}\Delta\tilde{\mathbf{R}}_{\mathrm{E}i\mathrm{E}k}\left(\tilde{v}_k\begin{bmatrix}\mathrm{S}_{\tilde{\omega}_k,\Delta t}\\ \mathrm{C}_{\tilde{\omega}_k,\Delta t}\\ 0\end{bmatrix}^{\wedge}\delta\boldsymbol{\phi}_{\mathrm{E}i\mathrm{E}k}-\boldsymbol{\varepsilon}_{k,\Delta t}^{\mathrm{p}}\right) \quad (22)$$
$$\doteq\Delta\tilde{\mathbf{p}}_{\mathrm{E}i\mathrm{E}j}-\delta\mathbf{p}_{\mathrm{E}i\mathrm{E}j}$$

where the encoder preintegrated rotation measurement, position measurement, and their noise from the estimation to the measurement are derived as follows.

$$\tilde{\theta}_{\mathrm{E}i\mathrm{E}j}=\tilde{\theta}_{\mathrm{E}i\mathrm{E}j-1}+\tilde{\theta}_{\mathrm{E}j-1\mathrm{E}j}=\sum_{k=i}^{j-1}\tilde{\omega}_k\Delta t,\Delta\tilde{\mathbf{R}}_{\mathrm{E}i\mathrm{E}j}=\mathrm{Exp}\left(\tilde{\boldsymbol{\theta}}_{\mathrm{E}i\mathrm{E}j}\right)$$
$$\Delta\tilde{\mathbf{p}}_{\mathrm{E}i\mathrm{E}j}=\Delta\tilde{\mathbf{p}}_{\mathrm{E}i\mathrm{E}j-1}+\frac{\tilde{v}_{j-1}}{\tilde{\omega}_{j-1}}\begin{bmatrix}\sin\tilde{\theta}_{\mathrm{E}i\mathrm{E}j}-\sin\tilde{\theta}_{\mathrm{E}i\mathrm{E}j-1}\\ \cos\tilde{\theta}_{\mathrm{E}i\mathrm{E}j-1}-\cos\tilde{\theta}_{\mathrm{E}i\mathrm{E}j}\\ 0\end{bmatrix}$$
$$\delta\boldsymbol{\phi}_{\mathrm{E}i\mathrm{E}j}\approx\sum_{k=i}^{j-1}\Delta\tilde{\mathbf{R}}_{\mathrm{E}k+1\mathrm{E}j}^{\mathrm{T}}\boldsymbol{\varepsilon}_{k,\Delta t}^{\mathrm{R}} \quad (23)$$
$$\delta\mathbf{p}_{\mathrm{E}i\mathrm{E}j}=\sum_{k=i}^{j-1}\Delta\tilde{\mathbf{R}}_{\mathrm{E}i\mathrm{E}k}\boldsymbol{\varepsilon}_{k,\Delta t}^{\mathrm{p}}+\mathbf{J}_{\delta\mathbf{p}_{\mathrm{E}i\mathrm{E}k+1}}^{\delta\boldsymbol{\phi}_{\mathrm{E}i\mathrm{E}k}}\delta\boldsymbol{\phi}_{\mathrm{E}i\mathrm{E}k}$$
$$\mathbf{J}_{\delta\mathbf{p}_{\mathrm{E}i\mathrm{E}k+1}}^{\delta\boldsymbol{\phi}_{\mathrm{E}i\mathrm{E}k}}=-\Delta\tilde{\mathbf{R}}_{\mathrm{E}i\mathrm{E}k}\tilde{v}_k\left[\mathrm{S}_{\tilde{\omega}_k,\Delta t}\quad \mathrm{C}_{\tilde{\omega}_k,\Delta t}\quad 0\right]^{\mathrm{T}\wedge}$$

where $\tilde{\boldsymbol{\theta}}_{\mathrm{E}i\mathrm{E}j}=[0\ 0\ \tilde{\theta}_{\mathrm{E}i\mathrm{E}j}]^{\mathrm{T}}$, and this exact solution is identical (5.9) in [22]. However, that book includes only the expression of the relative motion without using the proposed noise model in this paper.

By substituting (22) back into the definition of $\Delta\mathbf{R}_{\mathrm{E}i\mathrm{E}j}$ and $\Delta\mathbf{p}_{\mathrm{E}i\mathrm{E}j}$ in (21), we ultimately obtain our preintegrated encoder measurement model.

$$\Delta\tilde{\mathbf{R}}_{\mathrm{E}i\mathrm{E}j}\approx\mathbf{R}_{\mathrm{E}i\mathrm{E}j}\mathrm{Exp}\left(\delta\boldsymbol{\phi}_{\mathrm{E}i\mathrm{E}j}\right),\Delta\tilde{\mathbf{p}}_{\mathrm{E}i\mathrm{E}j}\approx\mathbf{p}_{\mathrm{E}i\mathrm{E}j}+\delta\mathbf{p}_{\mathrm{E}i\mathrm{E}j} \quad (24)$$

Through (23), we can easily find that noises $\delta\boldsymbol{\phi}_{\mathrm{E}i\mathrm{E}j}$ and $\delta\mathbf{p}_{\mathrm{E}i\mathrm{E}j}$ are zero mean and Gaussian since they are linear combinations of the zero-mean noise terms $\boldsymbol{\eta}_k^{(\cdot)\mathrm{d}}$ and $\delta\boldsymbol{\phi}_{\mathrm{E}i\mathrm{E}j}$.

## B. Residual Errors

Given (24), it is easy to write the encoder residual error (5) as $\mathbf{r}_{Eij}=\mathbf{e}_{\mathrm{E}}=[\mathbf{r}_{\Delta\mathrm{R}ij}{}^{\mathrm{T}}\ \mathbf{r}_{\Delta\mathrm{p}ij}{}^{\mathrm{T}}]^{\mathrm{T}}=[\mathbf{e}_{\mathrm{R}}{}^{\mathrm{T}}\ \mathbf{e}_{\mathrm{p}}{}^{\mathrm{T}}]^{\mathrm{T}}\in\mathbb{R}^6$.

$$\mathbf{e}_R = \mathrm{Log}\left(\Delta\tilde{\mathbf{R}}_{EiEj}^{T}\mathbf{R}_{EiEj}\right), \mathbf{e}_p = \mathbf{p}_{EiEj} - \Delta\tilde{\mathbf{p}}_{EiEj} \quad (25)$$

where the iterative form of the encoder preintegrated measurements is defined in (23). In addition, $\mathbf{p}_{EiEj}$ and $\mathbf{R}_{EiEj}$ can be obtained from $\mathbf{x}_i$ through prior calibrations $\mathbf{T}_{BC}$ and $\mathbf{T}_{CE}$. Here, Log($\mathbf{R}$)=$\theta\mathbf{a}$ restricts $\|\theta\| <= \pi$, where equality does not affect the error norm (5).

Then, we can also obtain the pose prediction (encoder) (Fig. 1) based on (25):

$$\hat{\mathbf{R}}_{EiEj} = \Delta\tilde{\mathbf{R}}_{EiEj}, \hat{\mathbf{p}}_{EiEj} = \Delta\tilde{\mathbf{p}}_{EiEj}, \mathbf{T}_{WCj} = \mathbf{T}_{WCi}\mathbf{T}_{CiCj},$$
$$\mathbf{T}_{CiCj} = \mathbf{T}_{CE}\hat{\mathbf{T}}_{EiEj}\mathbf{T}_{CE}^{-1}, \hat{\mathbf{T}}_{EiEj} = \begin{bmatrix} \hat{\mathbf{R}}_{EiEj} & \hat{\mathbf{p}}_{EiEj} \\ \mathbf{0}_{1\times 3} & 1 \end{bmatrix} \quad (26)$$

where $i$ and $j$ are just used as the consecutive indices of the frame as $j$=$i$+1.

We skip the section describing the noise propagation and Jacobians of the residual errors because they use similar techniques to those in [17] and yield the result in the appendix. Now, the key expression needed by the encoder residual error term to implement the code is proven.

## V. Materials and Methods

### A. Differential Wheeled Robot and Measurement equipment

As Fig. 4 shows, we designed a common differential wheeled robot equipped with one Kinect v2 RGB-D camera, two encoders that are, strictly speaking, Hall sensors installed in brushless motors, and one IMU sensor (Microstrain, 3DM-GX1). The robot has two back driving wheels driven by two 180 W brushless motors and two caster wheels as the front wheels. We used a robotic total station (Leica Geosystems, TS16) to measure the true value of the position (approximately 10 Hz) of a prism reflector fixed on the robot.

As Fig. 5 shows, the hierarchical control system hardware consists of a three-level system.

### B. Evaluation of SLAM

Since we could not find any public dataset containing encoder data, we developed our dataset including encoder information to evaluate the performance of our SLAM.

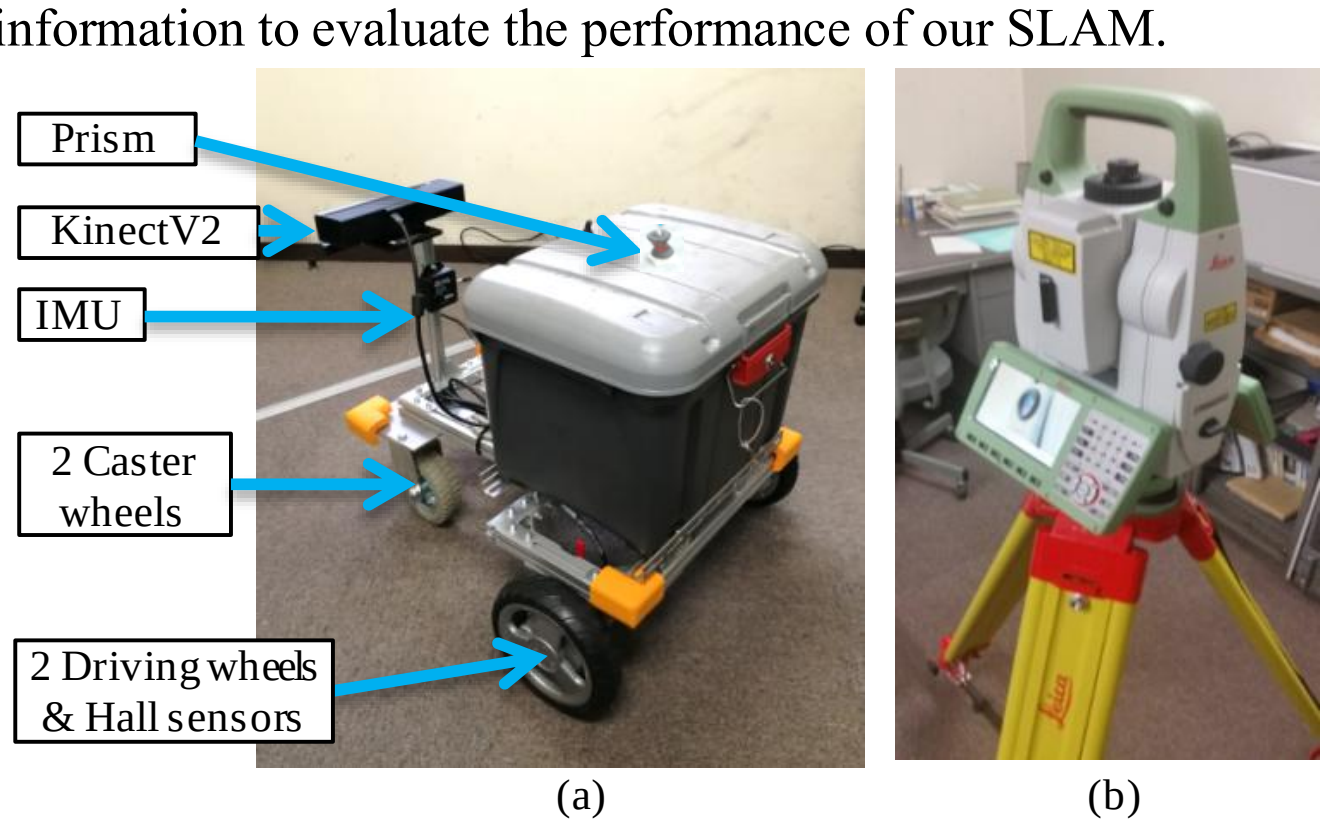

Fig. 4. Test equipment: (a) differential wheeled robot; (b) robotic total station.

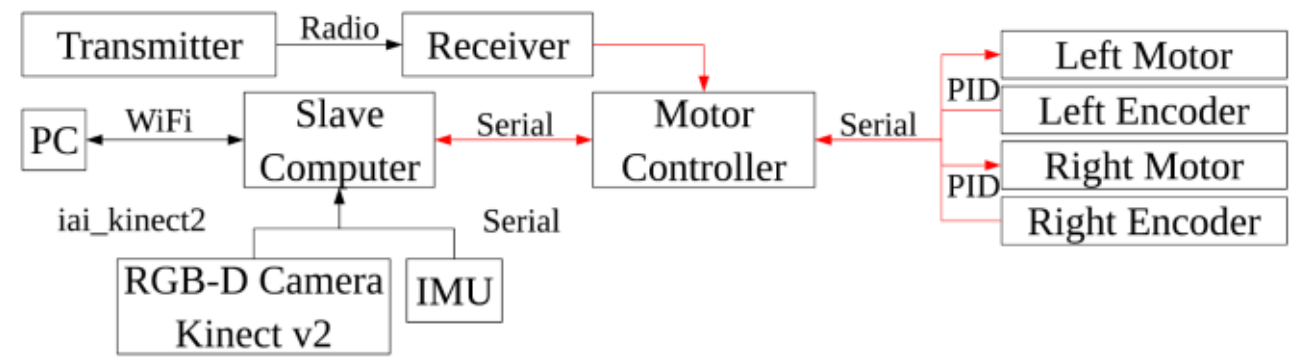

Fig. 5. Hierarchical control system hardware.

A public benchmark tool [8] was used to evaluate the absolute trajectory error (ATE), which shows the accuracy of the output of a real SLAM system by calculating the root mean squared error (RMSE). The estimated SLAM trajectory and ground truth trajectory, which are provided by a highly accurate system such as a total station whose accuracy can reach close to 1 mm [15], are the input files for the tool.

This tool can automatically align different world coordinate frames in the two input files by using the closed form method of Horn [8]. This method finds the rigid-body transformation S that corresponds to the least-square solution that maps the estimated trajectory onto the ground truth trajectory [8] $\mathbf{p}_{Wground_truth i}$=$\mathbf{S}\mathbf{p}_{Westimated i}$, where $\mathbf{S}$=$\mathbf{T}^{*}_{Wground_truthWestimated}$ and $\mathbf{p}$ here is homogeneous coordinate. Note that i and the rigid-body transformation here are the frames fixed on prism "P" and $\mathbf{S}$=$\mathbf{T}^{*}_{WLeicaW}$. Therefore, we transformed the estimated trajectory file from our RGB-D SLAM through $\mathbf{p}_{WPi}$ = $\mathbf{R}_{WCi}\mathbf{p}_{CP}$+$\mathbf{p}_{WCi}$, where $\mathbf{p}_{CP}$ can also be obtained by prior calibration.

## VI. Results and Discussion

We confirmed that our implementation of VIOS2 had similar accuracy to the VIORBSLAM paper on the EuRoC dataset, as shown in Table I. Note that the VIORBSLAM paper stated that the movement in V103 is too extreme for the monocular system to survive 15 s [16], but we found that sometimes it survives, and we provide the corresponding accuracy.

TABLE I
KeyFrame ATE(m) in EuRoC Dataset

| Sequence | VIOS2 Monocular | VIORBSLAM Monocular [16] | VIOS2 Stereo* | ORBSLAM2 Stereo* [9] |
|---|---|---|---|---|
| V101 | 0.019 | 0.023 | 0.036 | 0.035 |
| V102 | 0.025 | 0.027 | 0.016 | 0.020 |
| V103 | 0.049 | X | 0.020 | 0.048 |
| V201 | 0.014 | 0.018 | 0.023 | 0.037 |
| V202 | 0.024 | 0.024 | 0.010 | 0.035 |
| V203 | 0.031 | 0.047 | 0.065 | X |
| MH01 | 0.071 | 0.068 | 0.036 | 0.035 |
| MH02 | 0.057 | 0.073 | 0.017 | 0.018 |
| MH03 | 0.062 | 0.071 | 0.023 | 0.028 |
| MH04 | 0.068 | 0.087 | 0.053 | 0.119 |
| MH05 | 0.095 | 0.060 | 0.060 | 0.060 |

The full BA is executed at the end of all SLAM. These sequences record 30-Hz frames and 200-Hz IMU data. Stereo* means the result is compared to the post-processed ground truth files, and the Monocular values use the raw Vicon and Leica ground truth [16].

Additionally, we found that our VIOS2 stereo part could usually achieve better accuracy and stability compared to ORBSLAM2 [9], because the IMU information could help judge outliers and enable better prediction and restriction on poses. Especially in difficult sequences with faster motion, e.g.,

V103, V203, and MH04, the improvement through the introduction of the IMU sensor is clear. However, we should note that V203 will be lost for a small period due to severe motion blur and the lack of a strategy such as pure encoder edges, which can connect two states even if the visual part fails.

Our dataset sequences can be divided into three types: an indoor laboratory room, an indoor corridor, and a small outdoor farm. The speed and angular speed of the vehicle are controlled at approximately 0.4 m/s and 0.8 rad/s at maximum, respectively. We obtained the timestamps of the images (RGB and depth) from the open-source kinect2_bridge of iai_kinect2, which utilizes the Kinect v2 Linux driver named libfreenect2 and provides images to the robot operating system (ROS) [23]. The timestamps here are the receiving times of the images. Synchronization is performed by finding the two images with the closest times. Considering the different communication delays of USB 3.0 and the auto-changing exposure time of RGB images, which run at 15 fps under low light conditions, because depth images that use the time-of-flight (ToF) principle can stably run at 30 fps, the error of the synchronization between the RGB (color) and depth images is less than 1/30 s.

Here, we use two methods to record data: the low-frequency imwrite (OpenCV function) with ofstream (C++ function) direct polling method (Table II) and the high-frequency rosbag [23] and buffer vector method (Table III). In the first method, the timestamps are aligned with images through a polling technique, and the synchronization error of 4 sensors may be less than 1/15 s. The main sources of error are the image synchronization error, delay of serial communication with the IMU and encoders, and the registration process in the kinect2_bridge, which undistorts and aligns the pixels in the color and depth images. In the second method, rosbag has a faster storing speed than imwrite. Here, ofstream is only used in the end to store the IMU or encoder data vectors. The synchronization accuracy may be similar to the synchronization error of the images because the timestamps are the receiving time of the IMU/encoder data, and the serial communication delay is usually small (approximately 5 ms).

TABLE II
FRAME ATE IN THE (7 HZ/15 HZ) DATASET

| Sequence | RMSE (m) | Average Processing Time (ms) | Estimation Ratio (%) |
|---|---|---|---|
| | | Indoor Laboratory | |
| Encoders | 0.211 | 1 | 100 |
| ORBSLAM2 | 0.034 | 24 | 39 |
| VEOS2 | 0.047 | 26 | 100 |
| Sequence | | Indoor Corridor | |
| Encoders | 0.639 | 1 | 100 |
| ORBSLAM2 | 0.040 | 23 | 46 |
| VEOS2 | 0.069 | 24 | 100 |
| Sequence | | Small Outdoor Farm | |
| Encoders_CF | - | 1 | 100 |
| ORBSLAM2 | - | 48 | 21 |
| VEOS2 | - | 43 | 100 |

The full BA is executed at the end of all RGB-D SLAM. In addition, due to only the integration being calculated, the encoder method is very fast. These sequences record 7-Hz frames and timestamp aligned encoder data, while the minimum time interval is 1/15 s. Here, CF is the complementary filter. We attempted to simply fuse a low-frequency encoder and IMU quaternion information, which is output by 3DM-GX1 by the CF.

TABLE III
FRAME ATE IN THE (15 HZ, 200 HZ) DATASET

| Sequence | RMSE (m) | Average Processing Time (ms) | Estimation Ratio (%) |
|---|---|---|---|
| | | Indoor Laboratory | |
| Encoders | 0.090 | 1 | 100 |
| ORBSLAM2* | 0.024 | 27 | 71 |
| VIOS2 | 0.023 | 29 | 72 |
| VEOS2 | 0.026 | 27 | 100 |
| VIEOS2 | 0.024 | 30 | 100 |
| Sequence | | Indoor Corridor | |
| Encoders | 0.188 | 1 | 100 |
| ORBSLAM2 | 0.032 | 25 | 45 |
| VIOS2 | 0.031 | 29 | 34 |
| VEOS2 | 0.036 | 23 | 100 |
| VIEOS2 | 0.029 | 25 | 100 |

The full BA is executed at the end of all RGB-D SLAM. These sequences record 15-Hz frames and 200-Hz encoder + IMU data. ORBSLAM2* here is not stable; sometimes the estimation ratio could reach close to 90%, but the RMSE would be close to 0.08 m.

As Table II, Table III and Fig. 6 to Fig. 8 show, our RGB-D VEOS2 has much better accuracy than the pure encoder method, which only calculates the integration of the measured velocity and has an accuracy close to that of RGB-D ORBSLAM2 [9]. Although we did not measure the ground truth trajectory of the outdoor sequence, the robustness improvement could be easily evaluated from the estimation ratio. The accuracy improvement could only be qualitatively evaluated from the estimated trajectory of VEOS2 in comparison with the encoders (Fig. 6) because the ground truth trajectory should be a cycle, where the end position is near the start position. We offset -1 s (7/15 Hz), 0.1 s (15, 200 Hz) in the laboratory sequence and -0.5 s (7/15 Hz), 0.1 s (15, 200 Hz), in the corridor for the benchmark tool [8], which requires the manual alignment of the start time of two input trajectories. In addition, the estimation ratio is simply the ratio of the accumulated estimated time span to the total span of RGB images. The larger synchronization error and lower frequency in the 7/15 Hz datasets clearly lead to more error than in the 15- and 200-Hz datasets. Therefore, the requirement for measurement at high frequency is confirmed in real applications.

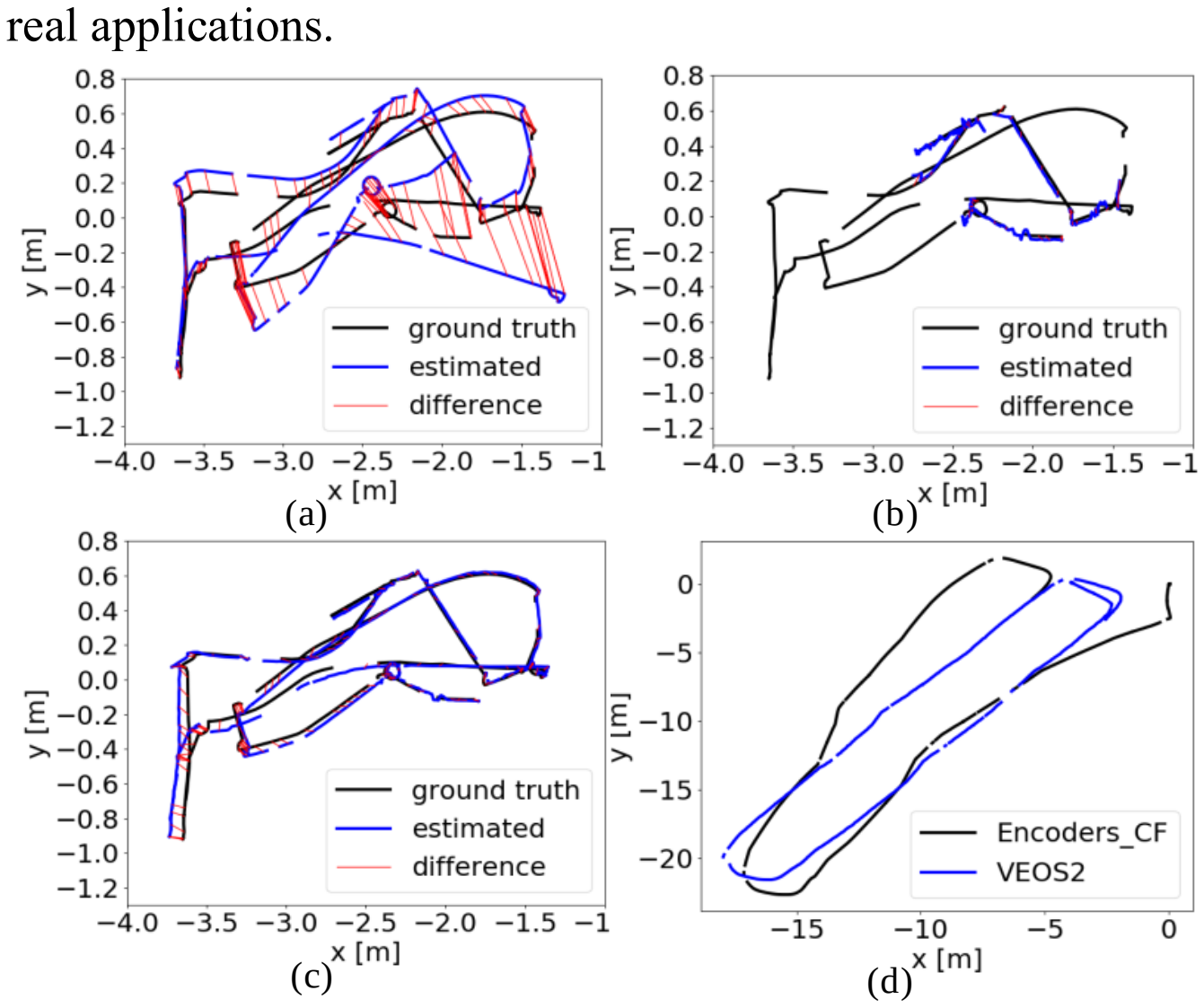

Fig. 6. Trajectories of the robot (7/15 Hz): (a) ATE of the encoder method in the laboratory, (b) ATE of ORBSLAM2 in the laboratory, (c)

ATE of VEOS2 in the laboratory, (d) absolute trajectory comparison of VEOS2 and encoders on the farm.

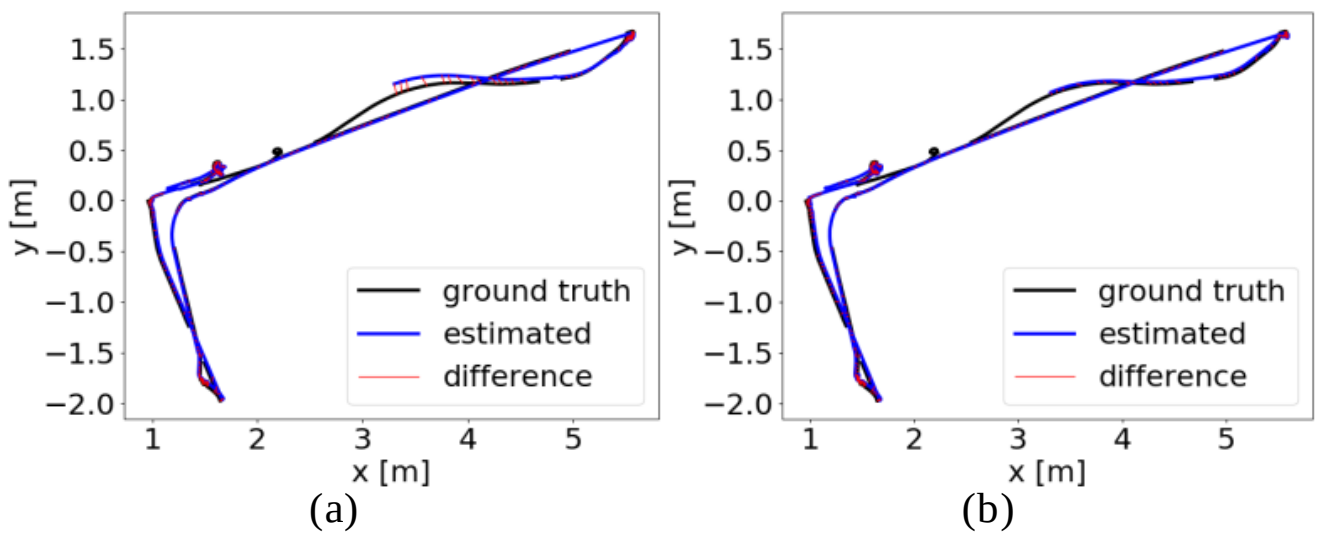

Fig. 7. Trajectories of the robot in the corridor (15, 200 Hz): (a) ATE of VEOS2, (b) ATE of VIEOS2.

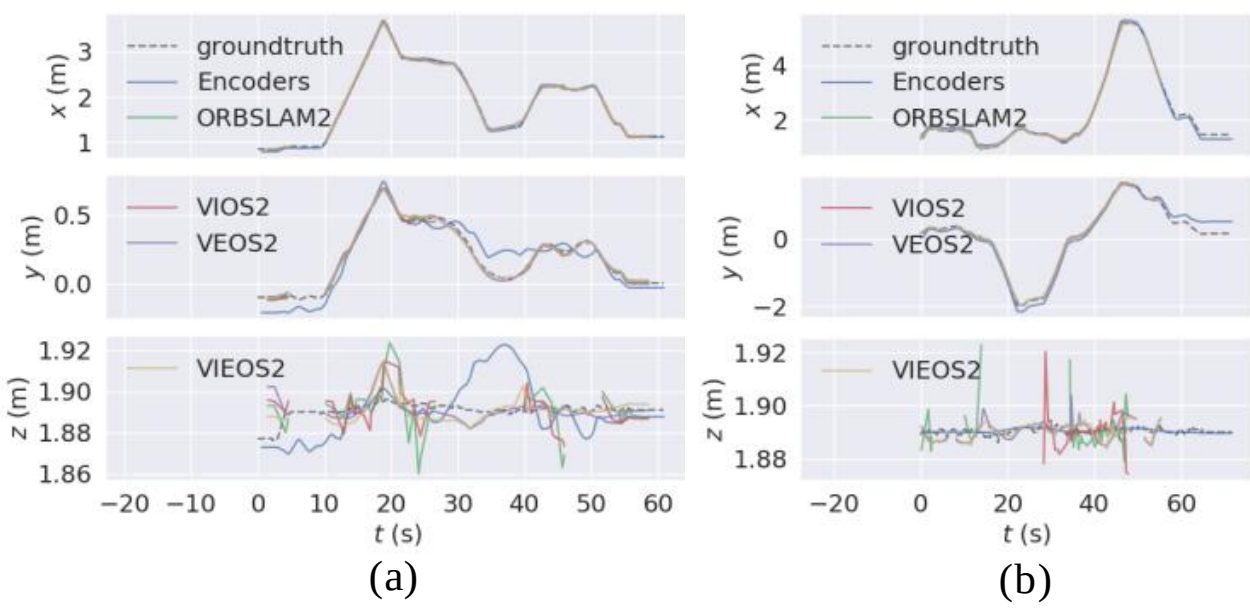

Fig. 8. Time-series graphs of ATE made by open-source evo [24] (15, 200 Hz): (a) laboratory, (b) corridor.

The failure in the laboratory room for ORBSLAM2, where the estimated trajectory is only approximately half of the whole ground truth, is caused mainly by the lack of feature points and fast motion. The failure in the corridor is additionally caused by the reflection light problem of ceiling lamps, and that of the small farm is mainly caused by the frame drop due to the unstable execution of the MCU. However, our VEOS2 and VIEOS2 can handle all of these temporary failure problems well and recover almost the entire trajectory with satisfactory accuracy compared to the part where RGB-D ORBSLAM2 can be used. Therefore, we conclude that our visual SLAM fused with extra odometry sensors has already achieved the main target of this paper.

Table III and Fig. 8 indicate that with a good calibration of the IMU covariance using the Allan variance [25], [20], VIOS2 can usually achieve a better accuracy than RGB-D ORBSLAM2 with a similar robustness, and VEOS2 or VIEOS2 can achieve a perfect robustness with a similar accuracy. VIOS2 or ORBSLAM2 usually has a slightly better accuracy than VEOS2 or VIEOS2, mainly because the latter two introduce pure encoder edges, which introduce both error and robustness. VIEOS2 can achieve a better accuracy than VEOS2, mainly because the pure encoder edges introduced in VIEOS2 can use extra IMU measurements. However, the encoder term is the key to achieving a very stable trajectory estimate on all of our datasets even at low frequency (Tables II). Therefore, VEOS2 and VIEOS2 perform better than the public ORBSLAM2, VIOS2, and encoders only. Our visual SLAM has a similar accuracy, but VEOS2 requires fewer sensors and is faster than VIEOS2 when an identical high estimation ratio is maintained. Ultimately, we prefer and recommend VEOS2 in the agricultural field.

The point cloud maps of the map points saved in these three sequences are shown in Fig. 9 using the point cloud library. For approximate navigation, the accuracy of the positions of obstacles, which means the points whose height is between the ground plane points and the height of our robot, are acceptable. However, some redundant parts can still be found due to the inevitable error introduced by pure encoder edges, which can be decreased by more loop closures and image measurements and better prior calibration. The point cloud map form of the map points is used mainly to display the visible error of the estimated trajectory of the keyframes. However, in ROS navigation, the grid map form that incorporates advanced filter techniques should be used for future research.

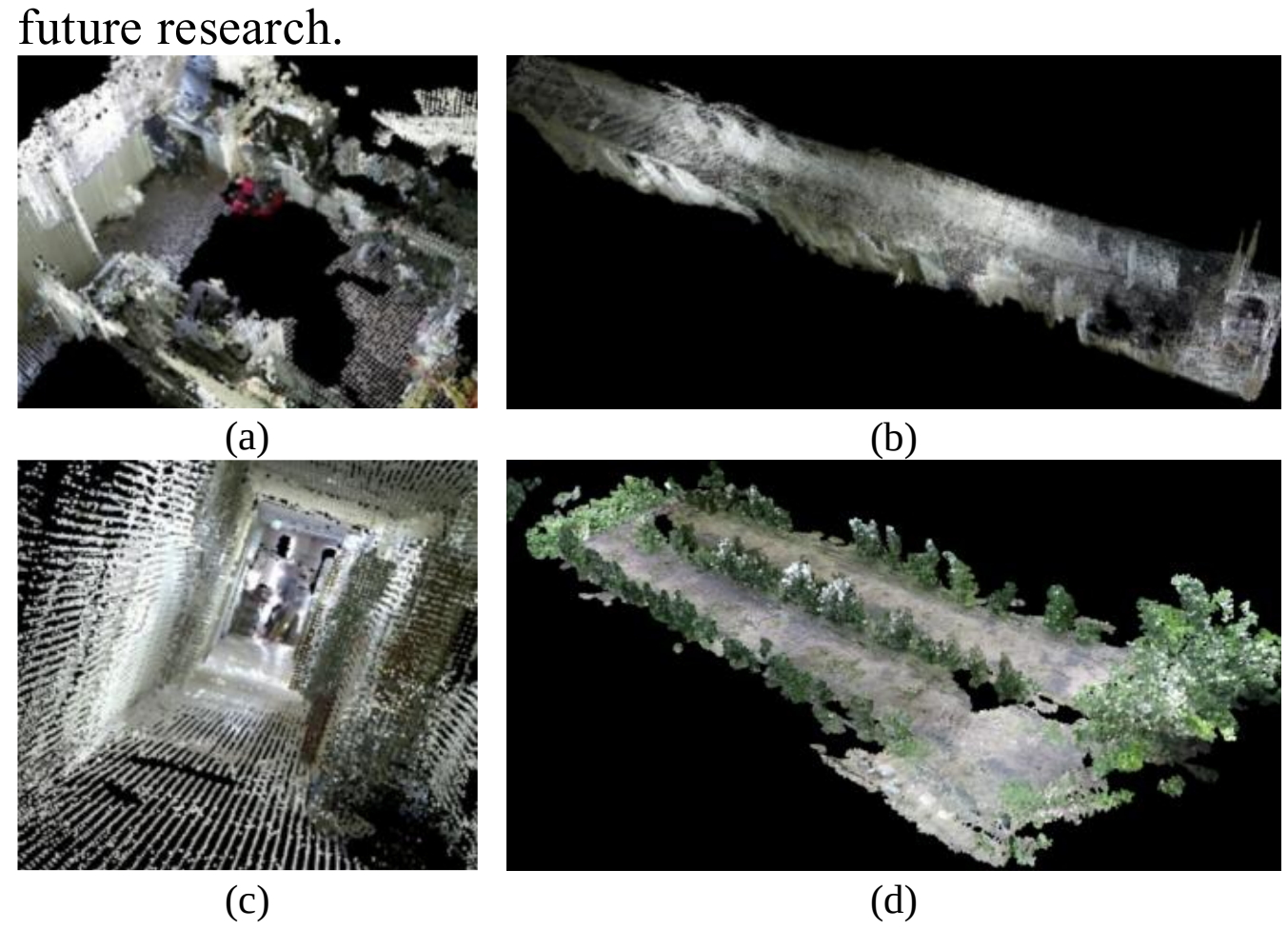

Fig. 9. Point cloud maps: (a) laboratory room, (b) corridor viewed from top, (c) corridor viewed from inside, (d) small farm.

## VII. Conclusion

We have proposed an open-source visual-inertial SLAM tightly coupled with encoders, which allows a novel trial of preintegration theory on different sensors compared to that in related works, e.g., [16], [17]. We used our dataset to verify that our SLAM maintained an accuracy similar to that of the state-of-the-art SLAM system, which is approximately 5 cm in the laboratory and corridor.

As a second contribution, we adjusted the flow chart of ORBSLAM2 [9] to show how to utilize the encoder information to help avoid the temporary failure problem of the pure visual SLAM when it is applied in the agricultural field and other settings. The experimental results confirm that our SLAM is more robust than the state-of-the-art SLAM system, e.g., our SLAM can predict 100% of a trajectory, whereas ORBSLAM2 or the one coupled with an IMU usually predicts less than half in the laboratory, corridor and farm.

We have also published a novel type of dataset that contains encoder information. For future research, the error terms of the encoder and IMU may be replaced by a complex error term through a Kalman filter to assume only flat planes during 2 frames instead of 2 keyframes. Alternatively, the interface of our SLAM with the ROS navigation node may be implemented to control our differential wheeled robot.

## APPENDIX

### A. Noise Propagation

To compute the preintegrated noise covariance, let us start from the preintegrated rotation noise in (23). By taking the last term ($k$=$j$-1) out of the sum and rearranging the terms as follows, we can write it in an iterative form.

$$\delta\boldsymbol{\phi}_{\mathrm{E}i\mathrm{E}j} \approx \Delta\tilde{\mathbf{R}}^{\mathrm{T}}_{\mathrm{E}j-1\mathrm{E}j}\delta\boldsymbol{\phi}_{\mathrm{E}i\mathrm{E}j-1} + \boldsymbol{\varepsilon}^{\mathrm{R}}_{j-1,\Delta t} \tag{27}$$

We repeat the same process for the preintegrated position noise.

$$\delta\mathbf{p}_{\mathrm{E}i\mathrm{E}j} = \delta\mathbf{p}_{\mathrm{E}i\mathrm{E}j-1} + \Delta\tilde{\mathbf{R}}_{\mathrm{E}i\mathrm{E}j-1}\boldsymbol{\varepsilon}^{\mathrm{p}}_{j-1,\Delta t} + \mathbf{J}^{\delta\boldsymbol{\phi}_{\mathrm{E}i\mathrm{E}j-1}}_{\delta\mathbf{p}_{\mathrm{E}i\mathrm{E}j}}\delta\boldsymbol{\phi}_{\mathrm{E}i\mathrm{E}j-1} \tag{28}$$

where $\mathbf{J}^{\delta\phi_{\mathrm{E}i\mathrm{E}k}}_{\delta\mathbf{p}_{\mathrm{E}i\mathrm{E}k+1}}$ is defined in (23), and $k$=$j$-1.

Recall that $\boldsymbol{\eta}_{Eij}$=$[\delta\boldsymbol{\phi}^{\mathrm{T}}_{\mathrm{E}i\mathrm{E}j}\ \delta\mathbf{p}^{\mathrm{T}}_{\mathrm{E}i\mathrm{E}j}]^{\mathrm{T}} \in \mathbb{R}^6$, and we have defined the encoder noise $\boldsymbol{\eta}^{\mathrm{ed}}$=$[\eta^{\mathrm{eld}}\ \eta^{\mathrm{erd}}]^{\mathrm{T}} \in \mathbb{R}^2$ with its covariance $\sum_{\mathrm{ed}}$ and encoder model noise $\boldsymbol{\eta}^{\mathrm{emd}}$=$[\eta^{\mathrm{mvdx}}\ \eta^{\mathrm{mvdy}}\ \eta^{\mathrm{mvdz}}\ \eta^{\mathrm{m\omega dx}}\ \eta^{\mathrm{m\omega dy}}\ \eta^{\mathrm{m\omega dz}}]^{\mathrm{T}} \in \mathbb{R}^6$ with its covariance $\sum_{\mathrm{emd}}$. Thus, we can finally write the former equations in compact matrix form as

$$\boldsymbol{\eta}_{Eij} = \mathbf{A}_{j-1}\boldsymbol{\eta}_{Eij-1} + \mathbf{B}_{j-1}\boldsymbol{\eta}^{\mathrm{ed}}_{j-1} + \mathbf{D}_{j-1}\boldsymbol{\eta}^{\mathrm{emd}}_{j-1} \tag{29}$$

Then, the iterative noise propagation of the covariance transfer can be computed as follows.

$$\boldsymbol{\Sigma}_{E_{ij}} = \mathbf{A}_{j-1}\boldsymbol{\Sigma}_{E_{ij-1}}\mathbf{A}^{\mathrm{T}}_{j-1} + \mathbf{B}_{j-1}\boldsymbol{\Sigma}_{\mathrm{ed}}\mathbf{B}^{\mathrm{T}}_{j-1} + \mathbf{D}_{j-1}\boldsymbol{\Sigma}_{\mathrm{emd}}\mathbf{D}^{\mathrm{T}}_{j-1} \tag{30}$$

starting from the initial conditions $\sum_{Eii}$=$\mathbf{0}_{6\times6}$. Here, matrices **A**, **B**, and **D** are derived as follows:

$$\mathbf{A}_{j-1} = \begin{bmatrix} \Delta\tilde{\mathbf{R}}^{\mathrm{T}}_{\mathrm{E}j-1\mathrm{E}j} & \mathbf{0}_{3\times3} \\ \mathbf{J}^{\delta\boldsymbol{\phi}_{\mathrm{E}i\mathrm{E}j-1}}_{\delta\mathbf{p}_{\mathrm{E}i\mathrm{E}j}} & \mathbf{I}_{3\times3} \end{bmatrix}, \mathbf{B}_{j-1} = \begin{bmatrix} \frac{1}{2r_c}\begin{bmatrix} \mathbf{0}_{2\times1} & \mathbf{0}_{2\times1} \\ -\Delta t & \Delta t \end{bmatrix} \\ \Delta\tilde{\mathbf{R}}_{\mathrm{E}i\mathrm{E}j-1}\mathbf{J}^{\boldsymbol{\eta}^{\mathrm{ed}}_{j-1}}_{\boldsymbol{\varepsilon}^{\mathrm{p}}_{j-1,\Delta t}} \end{bmatrix}, \Delta t = \Delta t_{j-1j}$$

$$\mathbf{J}^{\boldsymbol{\eta}^{\mathrm{ed}}_{j-1}}_{\boldsymbol{\varepsilon}^{\mathrm{p}}_{j-1,\Delta t}} = \frac{1}{2}\begin{bmatrix} \begin{bmatrix} \mathrm{S}_{\tilde{\omega}_{j-1},\Delta t} & \mathrm{C}_{\tilde{\omega}_{j-1},\Delta t} \end{bmatrix} - \mathbf{J}^{\eta^{\omega\mathrm{dz}}_{j-1}\ \mathrm{T}}_{\boldsymbol{\varepsilon}^{\mathrm{p}}_{j-1,\Delta t}} / r_c & 0 \\ \begin{bmatrix} \mathrm{S}_{\tilde{\omega}_{j-1},\Delta t} & \mathrm{C}_{\tilde{\omega}_{j-1},\Delta t} \end{bmatrix} + \mathbf{J}^{\eta^{\omega\mathrm{dz}}_{j-1}\ \mathrm{T}}_{\boldsymbol{\varepsilon}^{\mathrm{p}}_{j-1,\Delta t}} / r_c & 0 \end{bmatrix}^{\mathrm{T}}$$

$$\mathbf{J}^{\eta^{\omega\mathrm{dz}}_{j-1}}_{\boldsymbol{\varepsilon}^{\mathrm{p}}_{j-1,\Delta t}} = \frac{\tilde{v}_{j-1}}{\tilde{\omega}_{j-1}}\begin{bmatrix} \Delta t\cos\tilde{\theta}_{\mathrm{E}j-1\mathrm{E}j} - \mathrm{S}_{\tilde{\omega}_{j-1},\Delta t} & \Delta t\sin\tilde{\theta}_{\mathrm{E}j-1\mathrm{E}j} - \mathrm{C}_{\tilde{\omega}_{j-1},\Delta t} \end{bmatrix}^{\mathrm{T}}$$

$$\mathbf{D}_{j-1} = \begin{bmatrix} \begin{bmatrix} \mathbf{0}_{3\times3} & \Delta t\mathrm{J}_{\mathrm{r}}\left(\tilde{\boldsymbol{\omega}}_{j-1}\Delta t\right) \end{bmatrix} \\ \Delta\tilde{\mathbf{R}}_{\mathrm{E}i\mathrm{E}j-1}\begin{bmatrix} \mathbf{J}^{\boldsymbol{\eta}^{\mathrm{mvd}}_{j-1}}_{\boldsymbol{\varepsilon}^{\mathrm{p}}_{j-1,\Delta t}} & \mathbf{J}^{\boldsymbol{\eta}^{\mathrm{m\omega d}}_{j-1}}_{\boldsymbol{\varepsilon}^{\mathrm{p}}_{j-1,\Delta t}} \end{bmatrix} \end{bmatrix}, \mathbf{J}^{\boldsymbol{\eta}^{\mathrm{mvd}}_{j-1}}_{\boldsymbol{\varepsilon}^{\mathrm{p}}_{j-1,\Delta t}} = \Delta t\mathrm{J}_{\mathrm{r}}\left(-\tilde{\boldsymbol{\omega}}_{j-1}\Delta t\right) \tag{31}$$

$$\mathbf{J}^{\boldsymbol{\eta}^{\mathrm{m\omega d}}_{j-1}}_{\boldsymbol{\varepsilon}^{\mathrm{p}}_{j-1,\Delta t}} = \frac{\tilde{v}_{j-1}}{\tilde{\omega}_{j-1}}\begin{bmatrix} \mathbf{0}_{2\times1} & \mathbf{0}_{2\times1} & \mathbf{J}^{\eta^{\omega\mathrm{dz}}_{j-1}}_{\boldsymbol{\varepsilon}^{\mathrm{p}}_{j-1,\Delta t}} \\ \Delta t - \mathrm{S}_{\tilde{\omega}_{j-1},\Delta t} & -\mathrm{C}_{\tilde{\omega}_{j-1},\Delta t} & \end{bmatrix}$$

where $\tilde{\boldsymbol{\omega}}_{j-1}\Delta t$=[0,0, $\tilde{\omega}_{j-1}\Delta t$ ], and its simple right Jacobian $\mathrm{J}_{\mathrm{r}}$ is defined in (16).

### B. Jacobians of Residual Errors

Finally, the analytical expressions of the Jacobians of the encoder error term, which are used in g2o, can be computed as follows using a similar method [17].

The Jacobians in VIEOS2 can be obtained by (25):

$$\frac{\partial\mathbf{e}_{\mathrm{R}}}{\partial\delta\mathbf{p}_{\mathrm{WB}i}} = \frac{\partial\mathbf{e}_{\mathrm{R}}}{\partial\delta\mathbf{p}_{\mathrm{WB}j}} = \mathbf{0}_{3\times3}, \frac{\partial\mathbf{e}_{\mathrm{R}}}{\partial\delta\boldsymbol{\phi}_{\mathrm{WB}j}} = \mathrm{J}^{-1}_{\mathrm{r}}\left(\mathbf{e}_{\mathrm{R}}\right)\mathbf{R}^{\mathrm{T}}_{\mathrm{BE}} \tag{32}$$

$$\frac{\partial\mathbf{e}_{\mathrm{R}}}{\partial\delta\boldsymbol{\phi}_{\mathrm{WB}i}} = -\mathrm{J}^{-1}_{\mathrm{r}}\left(\mathbf{e}_{\mathrm{R}}\right)\left(\mathbf{R}^{\mathrm{T}}_{\mathrm{WB}i}\mathbf{R}_{\mathrm{WB}j}\mathbf{R}_{\mathrm{BE}}\right)^{\mathrm{T}}$$

$$\frac{\partial\mathbf{e}_{\mathrm{p}}}{\partial\delta\mathbf{p}_{\mathrm{WB}i}} = -\mathbf{R}^{\mathrm{T}}_{\mathrm{BE}}\mathbf{R}^{\mathrm{T}}_{\mathrm{WB}i}, \frac{\partial\mathbf{e}_{\mathrm{p}}}{\partial\delta\mathbf{p}_{\mathrm{WB}j}} = \mathbf{R}^{\mathrm{T}}_{\mathrm{BE}}\mathbf{R}^{\mathrm{T}}_{\mathrm{WB}i}$$

$$\frac{\partial\mathbf{e}_{\mathrm{p}}}{\partial\delta\boldsymbol{\phi}_{\mathrm{WB}i}} = \mathbf{R}^{\mathrm{T}}_{\mathrm{BE}}\left(\mathbf{R}^{\mathrm{T}}_{\mathrm{WB}i}\left(\mathbf{R}_{\mathrm{WB}j}\mathbf{p}_{\mathrm{BE}} + \mathbf{p}_{\mathrm{WB}j} - \mathbf{p}_{\mathrm{WB}i}\right)\right)^{\wedge}$$

$$\frac{\partial\mathbf{e}_{\mathrm{p}}}{\partial\delta\boldsymbol{\phi}_{\mathrm{WB}j}} = -\mathbf{R}^{\mathrm{T}}_{\mathrm{BE}}\mathbf{R}^{\mathrm{T}}_{\mathrm{WB}i}\mathbf{R}_{\mathrm{WB}j}\mathbf{p}_{\mathrm{BE}}{}^{\wedge}$$

The Jacobians in VEOS2 are:

$$\frac{\partial\mathbf{e}_{\mathrm{R}}}{\partial\delta\boldsymbol{\phi}_{\mathrm{WC}i}} = -\mathrm{J}^{-1}_{\mathrm{r}}\left(\mathbf{e}_{\mathrm{R}}\right)\left(\mathbf{R}^{\mathrm{T}}_{\mathrm{WC}i}\mathbf{R}_{\mathrm{WC}j}\mathbf{R}_{\mathrm{CE}}\right)^{\mathrm{T}} \tag{33}$$

$$\frac{\partial\mathbf{e}_{\mathrm{R}}}{\partial\delta\boldsymbol{\phi}_{\mathrm{WC}j}} = \mathrm{J}^{-1}_{\mathrm{r}}\left(\mathbf{e}_{\mathrm{R}}\right)\mathbf{R}^{\mathrm{T}}_{\mathrm{CE}}, \frac{\partial\mathbf{e}_{\mathrm{R}}}{\partial\delta\mathbf{p}_{\mathrm{WC}i}} = \frac{\partial\mathbf{e}_{\mathrm{R}}}{\partial\delta\mathbf{p}_{\mathrm{WC}j}} = \mathbf{0}_{3\times3}$$

$$\frac{\partial\mathbf{e}_{\mathrm{p}}}{\partial\delta\boldsymbol{\phi}_{\mathrm{WC}i}} = \mathbf{R}^{\mathrm{T}}_{\mathrm{CE}}\left(\mathbf{R}^{\mathrm{T}}_{\mathrm{WC}i}\left(\mathbf{R}_{\mathrm{WC}j}\mathbf{p}_{\mathrm{CE}} + \mathbf{p}_{\mathrm{WC}j} - \mathbf{p}_{\mathrm{WC}i}\right)\right)^{\wedge}$$

$$\frac{\partial\mathbf{e}_{\mathrm{p}}}{\partial\delta\boldsymbol{\phi}_{\mathrm{WC}j}} = -\mathbf{R}^{\mathrm{T}}_{\mathrm{CE}}\mathbf{R}^{\mathrm{T}}_{\mathrm{WC}i}\mathbf{R}_{\mathrm{WC}j}\mathbf{p}_{\mathrm{CE}}{}^{\wedge}$$

$$\frac{\partial\mathbf{e}_{\mathrm{p}}}{\partial\delta\mathbf{p}_{\mathrm{WC}i}} = -\mathbf{R}^{\mathrm{T}}_{\mathrm{CE}}\mathbf{R}^{\mathrm{T}}_{\mathrm{WC}i}, \frac{\partial\mathbf{e}_{\mathrm{p}}}{\partial\delta\mathbf{p}_{\mathrm{WC}j}} = \mathbf{R}^{\mathrm{T}}_{\mathrm{CE}}\mathbf{R}^{\mathrm{T}}_{\mathrm{WC}i}$$

## ACKNOWLEDGMENTS

The authors express their most sincere gratitude to Mr. Takeshi Yusa for his help in building the platform.

The first author thanks all members of the Laboratory of Biological and Mechanical Engineering for assistance over the past 2 years.

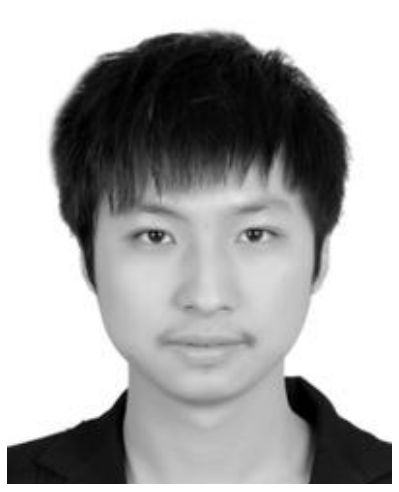

**Zhanghao Zhu** was born on Chongming Island, Shanghai, SH, China, in 1993. He received his B.S. degree in Mechanical Engineering and Automation from Shanghai Jiao Tong University in 2015 and his M.S. degree in Biological Engineering from the University of Tokyo in 2018.

From 2018 until 2021, he was an image processing and 3A algorithm engineer at Da-Jiang Innovations Science and Technology Co. (DJI). Since 2021, he has been a SLAM algorithm engineer at the YVR department of PLAY FOR DREAM.

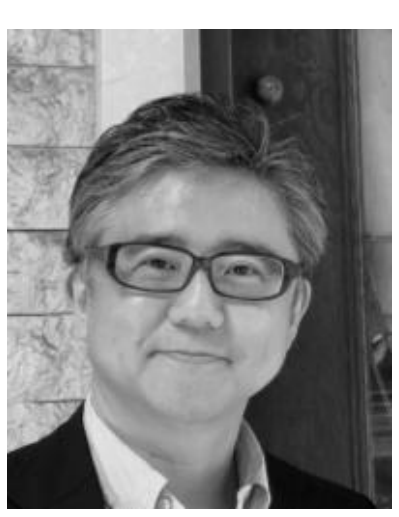

**Yutaka Kaizu** was born in Tokyo, Japan, in 1971. He received his B.S. degree in Agriculture from the University of Tokyo, Japan, in 1995, his M.S. degree in Agriculture from the Graduate School of Agricultural and Life Sciences, the University of Tokyo, Japan, in 1997 and his Ph.D. degree in Agriculture from the Graduate School of the University of Tokyo, Japan, in 2003.

From 1997 to 1999, he was an engineer at Kubota Corporation. From 1999 to 2006, he was an assistant professor at the University of Tokyo. From 2006 to 2012, he was an associate professor at Hokkaido University, Japan. Since 2012 , he has been an associate professor at the University of Tokyo.

His research interest is the development of agricultural robots that can autonomously move in the agricultural field.

**Kenichi Furuhashi** received his B.S. degree in Agriculture from the University of Tokyo, Japan, in 2011, and his M.S. and Ph.D. degrees in Agriculture from the Graduate School of Agricultural and Life Sciences, University of Tokyo, Japan, in 2013 and 2016, respectively. From 2016 to 2020, he was a researcher at the National Agriculture and Food Research Organization (NARO), Japan. Since 2020, he has been an assistant professor at the University of Tokyo, Japan. His fields are biomass energy and agricultural machinery.

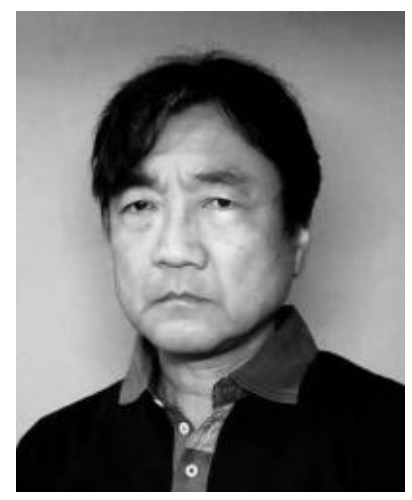

**Kenji Imou** received his B.S. degree in Agricultural Machinery from Mie University, Japan, in 1982, and his M.S. and Ph.D. degrees in Agricultural Engineering from the University of Tokyo, Japan, in 1984 and 1988, respectively. He worked in the fields of agricultural machinery automation and renewable energy. From 1986 to 1991, he was an Assistant Professor at the Faculty of Agriculture, the University of Tokyo. From 1991 to 1997, he was a Lecturer/Associate Professor at the Faculty of Agriculture, Utsunomiya University, Japan. Since 1997, he has been an Associate Professor/Professor at the Graduate School of Agricultural and Life Sciences, the University of Tokyo.

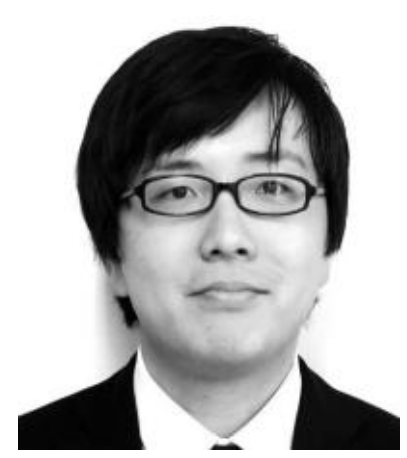